\RequirePackage{fix-cm}
\documentclass[smallextended]{svjour3}       % onecolumn (second format)
\smartqed  % flush right qed marks, e.g. at end of proof
\usepackage{graphicx}
\usepackage{todonotes}
\usepackage{tablefootnote}
\usepackage{lingmacros}
\usepackage{url}
\usepackage{multirow}
\usepackage{verbatim}
\begin{document}

\title{Predicting Lexical Complexity in English Texts: The Complex 2.0 Dataset}
% Grants or other notes about the article that should go on the front
% page should be placed within the \thanks{} command in the title
% (and the %-sign in front of \thanks{} should be deleted)
%
% General acknowledgments should be placed at the end of the article.

\subtitle{}


\author{Matthew Shardlow         \and
        Richard Evans \and 
        Marcos Zampieri %etc.
}


\institute{Matthew Shardlow  \at
              Manchester Metropolitan University, UK \\
              \email{m.shardlow@mmu.ac.uk}           %  \\
           \and
           Richard Evans\at
            University of Wolverhampton, UK \\
            \email{r.j.evans@wlv.ac.uk} 
              \and
           Marcos Zampieri \at
           Rochester Institute of Technology, USA \\
           \email{marcos.zampieri@rit.edu} 
}

\date{}

\maketitle

\begin{abstract}
Identifying words which may cause difficulty for a reader is an essential step in most lexical text simplification systems prior to lexical substitution and can also be used for assessing the readability of a text. This task is commonly referred to as Complex Word Identification (CWI) and is often modelled as a supervised classification problem. For training such systems, annotated datasets in which words and sometimes multi-word expressions are labelled regarding complexity are required. In this paper we analyze previous work carried out in this task and investigate the properties of CWI datasets for English. We develop a protocol for the annotation of lexical complexity and use this to annotate a new dataset, CompLex 2.0. We present experiments using both new and old datasets to investigate the nature of lexical complexity. We found that a Likert-scale annotation protocol provides an objective setting that is superior for identifying the complexity of words compared to a binary annotation protocol. We release a new dataset using our new protocol to promote the task of Lexical Complexity Prediction.
\end{abstract}

\section{Introduction}
\label{intro}

Predicting lexical complexity can enable systems to better guide a user to an appropriate text, or tailor it to their needs. The task of automatically identifying which words are likely to be considered complex by a given target population is known as Complex Word Identification (CWI) and it constitutes an important step in most lexical simplification pipelines \cite{paetzold2017survey}. 

The topic has gained significant attention in the last few years, particularly for English --- which is also the focus of our study. A number of studies have been published on predicting complexity of both single words and multi-word expressions (MWEs) including two recent competitions organized on the topic, CWI 2016 and CWI 2018, discussed in detail in Section \ref{sec:survey}. The first shared task on CWI was organized at SemEval in 2016 \cite{paetzold-specia:2016:SemEval1} providing participants with an English dataset in which words in context were annotated as non-complex (0) or complex (1) by a pool of human annotators. The goal was to predict this binary value for the target words in the test set. A post-competition analysis of the CWI 2016 results \cite{zampieri-EtAl:2017:NLPTEA} examined the performance of the participating systems and evidenced how challenging CWI 2016 was with respect to the distribution (more testing than training instances) and annotation type. 

%\todo[]{For newcomers, do we need to explain a bit more about what ``aggregated'' means in CWI annotation?} of its dataset. 

The second edition of the CWI shared task was organized in 2018 at the BEA workshop \cite{yimam2018report}. CWI 2018 featured multilingual (English, Spanish, German, and French) and multi-domain datasets \cite{yimam-EtAl:2017:RANLP}. Unlike in CWI 2016, predictions were evaluated not only in a binary classification setting but also in terms of probabilistic classification in which systems were asked to assign the probability of the given target word in its particular context being complex. Although CWI 2018 provided an element of regression, the continuous complexity value of each word was calculated as the proportion of annotators that found a word complex. For example, if 5 out of 10 annotators labeled a word as complex then the word was given a score of 0.5. This measure relies on an aggregation of absolute binary judgments of complexity to give a continuous value. 

Instead of using binary judgments, the CompLex dataset uses Likert Scale judgments \cite{shardlow-etal-2020-complex}, for which the specification is discussed in depth in Section \ref{sec:spec_cwi}. CompLex is a multi-domain English dataset annotated with a 5-point Likert scale (1-5) corresponding to the annotators comprehension and familiarity with the words in which 1 represents {\em very easy} and 5 represents {\em very difficult}. The CompLex dataset was used as the official dataset of SemEval-2021 Task 1: Lexical Complexity Prediction (LCP)\cite{shardlow2021semeval}. The goal of LCP 2021 is to predict this complexity score for each target word in context in the test set. 

%Marcos: @Matt, a bit more to the goals of the paper here?  
In this paper, we investigate properties of multiple annotated English lexical complexity datasets such as the aformentioned CWI datasets and others from the literature \cite{maddela-xu-2018-word}. We investigate the types of features that make words complex. We analyse the shortcomings of the previous CWI datasets and use this to motivate the specification of a new type of CWI dataset, focusing not on complex-word identification (CWI), but instead on lexical complexity prediction (LCP), that is CWI in a continuous-label setting. We further develop a dataset based on adding additional annotations to the existing CompLex 1.0 to create our new dataset, CompLex 2.0, and use this to provide experiments into the nature of lexical complexity.

The main contributions of this paper are:
\begin{itemize}
    \item A concise yet comprehensive survey of the two editions of the CWI shared tasks organized in 2016 and 2018;
    \item An investigation into the types of features that correlate with lexical complexity;
    \item A qualitative analysis of the CWI--2016 \cite{paetzold-specia:2016:SemEval1}, CWI--2018 \cite{yimam2018report} and Maddela--2018 \cite{maddela-xu-2018-word} datasets, highlighting issues with the annotation protocols that were used;
    \item The specification of a new annotation protocol for the CWI task;
    \item An implementation of our specification, describing the annotation of a new dataset for CWI (CompLex 1.0 and 2.0);
    \item Experiments comparing the features affecting lexical complexity in our dataset, as compared to others;
    \item Experiments using our dataset, demonstrating the effects of genre on CWI.
\end{itemize}
\noindent The remainder of this paper is organized as follows. Section \ref{sec:survey} provides an overview of the previous CWI shared tasks. Section \ref{sec:analysis_of_features} provides a preliminary investigation into the types of features that correlate with complexity labels in previous CWI datasets. Section \ref{sec:spec_cwi} firstly discusses the datasets that have previously been used for CWI, highlighting issues in their annotation protocols in Section \ref{sec:building_on}, and then proposes a new protocol for constructing CWI datasets in Section \ref{sec:specification}. Section \ref{sec:complex2} reports on the construction of a new dataset following the specification previously laid out. Section \ref{sec:analysis_of_existing} compares the annotations in our new dataset to those of previous datasets by developing a categorical annotation scheme. Section \ref{sec:complex_experiments} shows further experiments demonstrating how our new corpus can be used to investigate the nature of lexical complexity. Finally, a discussion of our main thesis and conclusions of our work are presented in Sections \ref{sec:discussion} and \ref{sec:conclusion} respectively.

We have previously published the CompLex 1.0 data as a workshop paper \cite{shardlow-etal-2020-complex}. The CompLex 2.0 data was also described in the SemEval task description paper \cite{shardlow2021semeval}. In this paper, we seek to build upon these prior works to give an in depth and rounded treatment to the lexical complexity problem.

\section{Related Work}
\label{sec:survey}

There have been various studies which have both created datasets and explored computational models for CWI, particularly focusing on English texts \cite{shardlow13,shardlow2013comparison,gooding-kochmar-2019-complex,finnimore-etal-2019-strong}. These studies have addressed CWI as a stand-alone task or as part of lexical simplification pipelines. 

Given the direct application of CWI to lexical simplification systems, where the goal is to decide whether or not a word needs to be substituted for a simpler one, the clear majority of studies have addressed CWI as a binary classification task. That said, there have been multiple studies analyzing the shortcomings of approaching CWI as a binary classification task. Some studies have studied the relationship between classification performance and dataset annotation in an attempt to estimate the theoretical upper boundary of binary CWI systems \cite{zampieri-EtAl:2017:NLPTEA} while others have investigated alternative ways to model the task. One study posed that comparative judgments are more consistent than binary classification for CWI \cite{gooding2019comparative}.  

CWI is of direct interest to those working in lexical simplification as it forms the first part of the lexical simplification pipeline \cite{devlin-1998}. Before a word can be simplified, a decision must be made as to whether or not that word requires simplification. Simplification systems \cite{biran-2011,bott-2012}, then generate potential candidates for simplification and use a similar process to CWI to select the most simple candidate \cite{Paetzold2017}.

Comparative complexity is a related but distinct task to Lexical Complexity Prediction. In this task, two words are taken and a judgment is given to determine which is the most complex. A recent study found that annotations for comparative complexity were more consistent than binary classification \cite{gooding2019comparative}. Nonetheless, we have not focussed on comparative complexity in this work, but rather on continuous complexity. We are most interested in  the complexity of a word in it's original context, rather than in relation to another word.

The increased interest from the research community in CWI was the primary motivation for the organisation of the two editions of the aforementioned CWI shared task in 2016 and 2018. These shared tasks have made important benchmark datasets available to the community that are widely used beyond these competitions. In the next sub-sections we provide an overview of these two editions: CWI--2016 organized at SemEval 2016 \cite{paetzold-specia:2016:SemEval1} and CWI--2018 organized at the BEA workshop in 2018 \cite{yimam2018report}. We describe the task setup, present the datasets, and briefly discuss the approaches submitted by participants in the two editions of the competition. We also present the approaches and the features used by each system. Finally, we analyze the results obtained by the participants and the main challenges of each edition of the CWI Shared Task.

\subsection{CWI--2016}
\label{sec:CWI2016}

The first shared task on CWI was organized as Task 11 at the International Workshop on Semantic Evaluation (SemEval) in 2016.\footnote{\url{http://alt.qcri.org/semeval2016/task11/}} CWI--2016 provided participants with a manually annotated dataset in which words in context were labeled as complex or non-complex, where complexity is interpreted as whether a word was understood or not by a pool of 400 non-native speakers of English. CWI--2016 was therefore modelled as a binary text classification task at the word level. Participants were required to build systems to predict lexical complexity in sentences of the unlabeled test set and assign label 0 to non-complex words and 1 to complex ones. Two examples from the CWI--2016 dataset are shown below:

\enumsentence{A {\bf frenulum} is a small fold of tissue that secures or {\bf restricts} the {\bf motion} of a mobile organ in the body.}
\enumsentence{The name `kangaroo mouse' refers to the species' {\bf extraordinary} jumping ability, as well as its habit of {\bf bipedal} {\bf locomotion}.}

\noindent The words in bold: {\em frenulum, restricts,} and {\em motion} in Example 1, and {\em extraordinary, bipedal,} and {\em locomotion} in Example 2 were annotated by at least one of the annotators as complex and thus they were labeled as such in the training set. Adjacent words like {\em bipedal locomotion} do not represent multi-word expressions (MWEs) as they were annotated in isolation because the task set-up of CWI--2016 only considered single word annotations. Whilst MWEs were not considered in CWI--2016, they were studied in CWI--2018 (see Section \ref{sec:CWI2018}).

The dataset provided by the organizers of CWI--2016 contained a training set of 2,237 target words in 200 sentences. The training set was annotated by 20 annotators and a word was considered complex in the training set if at least one of the 20 annotators assigned it as so. The test set included 88,221 target words in 9,000 sentences and each word was annotated by only one annotator. Therefore, the ground truth label for each word in the test was attributed based on a single complexity judgement. According to the organisers of CWI--2016, this setup was devised to imitate a realistic scenario where the goal was to predict the individual needs of a speaker based on the needs of the target group  \cite{paetzold-specia:2016:SemEval1}. Finally, the data included in the CWI--2016 dataset comes from various sources such as the CW Corpus \cite{shardlow2013comparison}, the LexMTurk Corpus \cite{horn14}, and Simple Wikipedia \cite{kauchak13}. 

%\todo[]{I have the impression it was a frequency-based feature motivated by Zipf's findings. I'm not sure ``Zipfian frequency distribution'' is a feature.} 
CWI--2016 attracted a large number of participants. A total of 21 teams submitted 42 systems to the competition. A wide range of features such as word embeddings, word and character n-grams, word frequency, Zipfian frequency-based features, word length, morphological, syntactic, semantic, and psycholinguistic features were used by participants. A number of different approaches to classification were tested, ranging from traditional machine learning classifiers such as support vector machines (SVM), decision trees, random forest, and maximum entropy classifiers to deep learning classifiers, such as recurrent neural networks. In Table \ref{tab:approaches}, we list the approaches submitted to CWI--2016 by the 19 teams who wrote system description papers presented at SemEval. 

\begin{table*}[!ht]
\centering
\scalebox{0.95}{
  \begin{tabular}{lp{3.2cm}p{4.8cm}c}
\hline
  \bf Team & \bf Classifiers & \bf Features & \bf Paper \\ \hline
  
  \bf AI-KU & SVM & word embeddings of the target and surrounding words & \cite{kuru:2016:SemEval} \\
  
  \bf  Amrita-CEN & SVM & word embeddings and various semantic and morphological features & \cite{sp-kumar-kp:2016:SemEval} \\
    
\bf     BHASHA & SVM, Decision Tree & lexical and morphological features & \cite{choubey-pateria:2016:SemEval} \\
  
 \bf ClacEDLK & Random Forests & semantic, morphological, and psycholinguistic features & \cite{davoodi-kosseim:2016:SemEval}\\
  
 \bf  CoastalCPH & Neural Network, Logistic Regression & word frequencies and word embeddings & \cite{bingel-schluter-martinezalonso:2016:SemEval} \\
  
\bf    HMC & Decision Tree & lexical, semantic, syntactic and psycholinguistic features & \cite{quijada-medero:2016:SemEval} \\
    
  \bf  IIIT & Nearest Centroid & semantic and morphological features & \cite{palakurthi-mamidi:2016:SemEval} \\
    
    \bf JUNLP & Random Forest, Naive Bayes & semantic, lexicon-based, morphological and syntactic features & \cite{mukherjee-EtAl:2016:SemEval} \\
     
\bf     LTG & Decision Tree & n-grams and word length & \cite{malmasi-dras-zampieri:2016:SemEval} \\

  \bf    MACSAAR & Random Forest, SVM & Zipfian frequency distribution, word length & \cite{zampieri-tan-vangenabith:2016:SemEval} \\
    
\bf  MAZA & Meta-classifier & n-grams, word probability, word length & \cite{malmasi-zampieri:2016:SemEval} \\
  
 \bf   Melbourne & Weighted Random Forests & lexical and semantic features & \cite{brooke-uitdenbogerd-baldwin:2016:SemEval} \\
  
 \bf   PLUJAGH & Threshold-based methods & features extracted from Simple Wikipedia & \cite{wrobel:2016:SemEval} \\
  
 \bf   Pomona & Threshold-based methods & word frequencies & \cite{kauchak:2016:SemEval} \\
  
 \bf   Sensible & Ensemble Recurrent Neural Networks & word embeddings & \cite{nat:2016:SemEval} \\
  
 \bf SV000gg & System voting with threshold & morphological, lexical, and semantic features & \cite{paetzold-specia:2016:SemEval2} \\
  
\bf  TALN & Random Forest & lexical, morphological, semantic, and syntactic features & \cite{ronzano-EtAl:2016:SemEval} \\
  
\bf    USAAR & Bayesian Ridge classifiers & hand-crafted word sense entropy metric and language model perplexity & \cite{martinezmartinez-tan:2016:SemEval} \\
  
\bf  UWB & Maximum Entropy & word occurrence counts on Wikipedia documents & \cite{konkol:2016:SemEval} \\

  \hline
  \end{tabular}
}
\caption{Systems submitted to the CWI--2016 in alphabetical order. We include team names and a brief description of each system including features and classifiers used. A reference to each system description paper is provided for more information.}
\label{tab:approaches}
\end{table*}

In terms of performance the top-3 systems were team PLUJAGH \cite{wrobel:2016:SemEval}, LTG \cite{malmasi-dras-zampieri:2016:SemEval}, and MAZA \cite{malmasi-zampieri:2016:SemEval} which obtained 0.353, 0.312, and 0.308 F1-score respectively. The three teams used rather simple probabilistic models trained on features such as n-grams, word frequency, word length, and the presence of words in vocabulary lists extracted from Simple Wikipedia, introduced by PLUJAGH. The relatively low performance obtained by all teams, including the top-3 systems, evidences how challenging the CWI--2016 shared task was. Both the data annotation protocol and the training/test split, where 40 times more testing data than training data is available, contributed to making CWI--2016 a difficult task.  

A post-competition analysis was carried out using the output of all 42 systems submitted to CWI--2016 \cite{zampieri-EtAl:2017:NLPTEA}. Each system output to each test instance was used as a vote to build two ensemble models. The ensemble models were built using plurality voting which assigns the highest number of votes as the label of a given instance, and exploits an oracle which assigns the correct label for an instance if at least one of the systems predicted the ground truth label for that instance. The plurality vote serves to better understand the performance of the systems using the same dataset while the oracle is used to quantify the theoretical upper limit performance on the dataset \cite{kuncheva2001decision}. The study showed that the potential upper limit for the CWI--2016 dataset considering the output of the participating systems is 0.60 F1 score for the complex word class. The outcome confirms that the low performance of the systems is related to the way the data has been annotated. Finally, this study also confirmed the relationship between word length and lexical complexity annotation in this dataset, a feature used by many of the teams participating in CWI--2016 as well as in our present work.

\subsection{CWI--2018}
\label{sec:CWI2018}

Following the success of CWI--2016, the second edition, CWI--2018, was organized at the Workshop on the Innovative Use of NLP for Building Educational Applications (BEA) in 2018.\footnote{https://sites.google.com/view/cwisharedtask2018/} Unlike CWI--2016 which focused only on English, CWI--2018 featured English, French, German, and Spanish datasets opening new perspectives in research in this area. 

A total of four tracks were available at CWI--2018: English, German, and Spanish, in which training and testing data was available for each language, and French. The organizers released a French test set with no corresponding training set with the goal of deriving models for French CWI from the English, Spanish, and German datasets. CWI--2018 featured two sub-tasks: (i) a binary classification task similar to CWI--2016 where participants were asked to label the given target word in a particular context as complex or simple; (ii) a probabilistic classification task where participants were asked to give a probability of the given target word in a particular context being complex. 

In terms of data, CWI--2018 used the \emph{CWIG3G2} dataset \cite{yimam-EtAl:2017:RANLP} in English, German, and Spanish. The English dataset contains texts from three domains, \emph{News}, \emph{WikiNews}, and \emph{Wikipedia} articles and the evaluation was carried out per domain. To allow cross-lingual learning, a dataset for French was collected using the same methodology as the one used for the CWIG3G2 corpus. Another important difference between CWI--2016 and CWI--2018 is that the \emph{CWIG3G2} featured annotation of both single words and MWEs while the dataset used in CWI--2016 only considered single words.

In terms of participation, CWI--2018 attracted 12 teams in different task/track combinations. In Table \ref{tab:approaches2018}, we list the approaches submitted to the English binary classification single word track by the 10 teams who wrote system description papers presented at BEA. Most teams tried multiple approaches and here we describe the teams' best-performing ones according to their system description papers.

\begin{table*}[!ht]
\centering
\scalebox{0.95}{
  \begin{tabular}{lp{3cm}p{4.5cm}c}
\hline
  \bf Team & \bf Classifiers & \bf Features & \bf Paper \\ \hline

\bf   CAMB & Adaboost & N-grams, WordNet features, POS tags, dependency parsing relations, psycholinguistic features. & \cite{syspaper7} \\

\bf   CFILT\_IITB & Voting ensemble & Word length, syllable counts, vowel counts, WordNet-based features. & \cite{syspaper10} \\

\bf   hu-berlin & Naive Bayes & Character n-grams & \cite{syspaper3} \\

\bf   ITEC & LSTM & Word length, word and character embeddings, frequency count, psycholinguistics features. & \cite{syspaper6}\\

\bf   LaSTUS/TALN & SVM, Random Forest & Word length, word embeddings, semantic and contextual features.  & \cite{syspaper9} \\

\bf   NILC &  XGBoost & N-grams, word length, number of syllables, WordNet-based features. & \cite{syspaper4} \\

\bf   NLP-CIC & Tree Ensembles and CNNs & Word frequency, syntactic and lexical features, psycholinguistic features, and word embeddings.  &  \cite{syspaper11} \\

\bf   SB@GU & Extra Trees & Word length, number of syllables, n-grams, frequency distribution. & \cite{syspaper2} \\

\bf   TMU & Random Forest & Word length, word frequency, probability features derived from corpora. &\cite{syspaper5} \\

\bf  UnibucKernel & Kernel-based learning with SVMs. & Character n-grams, semantic features, and word embeddings. & \cite{syspaper1} \\

  \hline
  \end{tabular}
}
\caption{Systems submitted to the CWI--2018 English binary classification single word track. We include team names and a brief description of each system including features and classifiers used. A reference to each system description paper is provided for more information.}
\label{tab:approaches2018}
\end{table*}

For the English binary classification single word track, the organizers reported the performance by all teams per domain. Team CAMB obtained the best performance for the three domains: 0.8736 F1-score on News, 0.8400 F1-score on WikiNews, and 0.8115 F1-score on Wikipedia. We observed that for all teams the performance on the News domain was generally substantially higher than the performance obtained in the two other domains. Several teams used the opportunity to compare multiple approaches for this task and many of them reported that traditional machine learning classifiers were more accurate than deep neural networks \cite{syspaper4,syspaper2}.

\section{Analysis of Features of Complex Words}\label{sec:analysis_of_features}

%Our inspection of words that are judged to be complex by a majority of annotators in the CWI--2016 and CWI--2018 datasets pointed to several intuitive explanations:

Upon analysing the datasets and system features used in CWI--2016 and CWI--2018, we noticed several intuitive explanations as to why a word may be judged as complex, or not:
\begin{itemize}
    \item The word is archaic.
    \item The word is a borrowing from another language or refers to a concept that is atypical in the culture of the reader.
    \item The word is uncommon and many people are not generally exposed to it.
    \item The word refers to a very specialised concept.
    \item Although the word is common, it is being used with an uncommon meaning in the given context.
\end{itemize}
\noindent These possible characteristics  motivated us to represent input words as sets of indicative linguistic features for the purpose of CWI. We used 378 features to represent words in our data set. These include psycholinguistic features derived from the MRC database \cite{wilson-1988}, word embeddings, and several other features with the potential to capture our intuitions about lexical complexity.

% \todo[inline,backgroundcolor=white, size=\small]{Another possible issue might relate to the fact that the word resembles a word in another language whose meaning the annotator knows, but can't be sure it means exactly the same thing. Or perhaps the resemblance to a word in another language is offset by the strangeness of its context.}

% \todo[inline,backgroundcolor=white, size=\small]{TO BE DELETED: in the subsequent annotation task, could we ask annotators both to judge complexity and fill out a simple opinion item about each word to ask why they thought it was complex (if they did). e.g. I know the word but I can't quite remember what it means; I've no idea what this means; it looks an old fashioned word; it doesn't look like an English word; the word is so morphologically complex that I can't decide whether it means one thing or the opposite of that thing; I know this word but it's being used in a strange way here.}

% \todo[inline,backgroundcolor=white, size=\small]{Talk about the features used to capture the intuitions: archaic word dictionary, information about word etymology (I think this information was not exploited in end because we need an API for wikitionary which has detailed etymological information; information about Wikipedia infoboxes, which tend to occur in pages about specialised concepts. We used WSD in an attempt to get at the semantic information and also contextual word embeddings.}

Values of the psycholinguistic features of words were obtained using the API to the MRC database. Many of the resources included in the database were built before 1998. These were derived through rigorous psycholinguistic testing, and as a result are of restricted size (offering relatively poor coverage of current English vocabulary). For this reason, in addition to specifying the values of these features directly from the database, we included binary features to indicate whether or not the word occurs in the MRC database.

We used information about whether or not the Wikipedia entry for the word includes an infobox element to indicate its degree of specialisation. Wiki-pedia\footnote{\url{https://en.wikipedia.org/wiki/Help:Infobox}. Last accessed 16th September 2021} describes infoboxes as:

\begin{quotation}
``[...] a fixed-format table usually added to the top right-hand corner of articles to consistently present a summary of some unifying aspect that the articles share and sometimes to improve navigation to other interrelated articles. Many infoboxes also emit structured metadata which is sourced by DBpedia and other third party re-users. The generalized infobox feature grew out of the original taxoboxes (taxonomy infoboxes) that editors developed to visually express the scientific classification of organisms.''
\end{quotation}

We observed that entries for specialised vocabulary (e.g. \emph{Gharial}) frequently contain infobox elements of various types (e.g. \emph{biota}). We extracted features encoding information about the occurrence and type of infobox element as an indicator of the level of specialisation of the word. We view this as a type of coarse-grained semantic information which is available for a relatively large proportion of words: more than 76\% of those occurring in the CWI-2016 and CWI-2018 datasets.  

% \todo[]{We could include a screenshot from Wikipedia here? Check Gharial: https://en.wikipedia.org/wiki/Gharial infobox biota. Problem is the Gharial infobox is too large.}

%\todo[]{I suggest replacing ``prediction models'' by the most appropriate one of ``lexical complexity prediction models'' and ``models for complex word identification''.}
The full feature set is displayed in tables \ref{table:wordFeaturesAJ} and \ref{table:wordFeaturesKT}. Given that it encodes well-motivated psycholinguistic information and includes features which capture our intuitions about lexical complexity, we consider this feature set to be suitable for use in the derivation of models for CWI. We processed the human-annotated CWI--2016 and CWI--2018 datasets to represent words as feature vectors using the features in these tables. 

% First, a quick comment about the features evaluated using the API to the MRC database. Many of the resources included in the database were built before 1988 and some of them exploit information from the 1960s. It is therefore built from datasets of relatively restricted size. As a result, they may offer poor coverage of current English vocabulary, especially with respect to specialised words. 

% \todo[inline,backgroundcolor=white,size=\small]{Note that many of the features are paired with features to indicate whether or not they are listed in the MRC resources. i.e the word is in the word list vs. the word is not in the word list.}

\begin{table}[ht]
% \begin{footnotesize}
\begin{small}
\begin{centering}
% \resizebox{\textwidth}{!}{
\begin{tabular}{llll}
 \hline
\bf ID & \textbf{Feature} & \textbf{Type} & \textbf{Definition} \\ \hline
 A & \parbox{2cm}{Frequent} & Binary & \parbox[l]{6cm}{One of the $10\,000$ most frequent words listed in Wiktionary} \\
  &  & & \parbox[l]{6cm}{} \\
 B & Archaic & Binary & \parbox[l]{6cm}{Listed in an archaic word list.$^\dagger$} \\
  &  & & \parbox[l]{6cm}{} \\
 C & \parbox{2cm}{Length (normalised)} & Numerical & \parbox[l]{6cm}{Length of the word divided by 50.$^\ddag$} \\
  &  & & \parbox[l]{6cm}{} \\
 D & Plurality & Binary & \parbox[l]{6cm}{5 features indicating whether the word is plural, has no plural form, is a singular form, is both singular and plural form, or is plural but acts singular.} \\
  &  & & \parbox[l]{6cm}{} \\
 E & Familiarity & \parbox[r]{1.5cm}{Numerical (100-700)} & \parbox[l]{6cm}{Familiarity score, derived by merging three sets of norms: Paivio (unpublished; these are an expansion of the norms of Paivio, Yuille, and Madigan \cite{paivio-1968}), Toglia and Battig \cite{toglia-1978}, and Gilhooly and Logie \cite{gilhooly-1980}). See Wilson \cite{wilson-1988} for more details on these metrics.} \\
  &  & & \parbox[l]{6cm}{} \\
 F & Concreteness & \parbox[r]{1.5cm}{Numerical (100-700)} & \parbox[l]{6cm}{Concreteness score, listed in the MRC Database} \\
  &  & & \parbox[l]{6cm}{} \\
 G & Imageability & \parbox[r]{1.5cm}{Numerical (100-700)} & \parbox[l]{6cm}{Imageability score of the word, listed in the MRC Database} \\
  &  & & \parbox[l]{6cm}{} \\
H & Brown & Numerical & \parbox[l]{6cm}{Frequency count of the word in the London-Lund Corpus of English Conversation \cite{svartvik-1980}} \\
 &  & & \parbox[l]{6cm}{} \\
I & KF$_{FREQ}$ & Numerical & \parbox[l]{6cm}{Frequency count of the word in the Ku\v{c}era and Francis \cite{kucera-1967} frequency list, derived from the Brown corpus.} \\
 &  & & \parbox[l]{6cm}{} \\
J & TL$_{FREQ}$ & Numerical & \parbox[l]{6cm}{Frequency listed in Thorndike and Lorge’s \cite{thorndike-1944} L count, which combines the counts of morphological variants of the word in a reference corpus.} \\
\hline
\end{tabular}
\end{centering}
\end{small}
\caption{Features (A-J) used to represent words. $\dag$ Available at https://archive.org/stream/dictionaryofarch028421mbp/dictionaryofarch~028421mbp\_djvu.txt. Last accessed 26th February 2019. $\ddag$ Longest word in English being 45 characters (pneumonoultramicroscopicsilicovolcanoconiosis).} \label{table:wordFeaturesAJ}
\end{table}

\begin{table}[ht]
% \begin{footnotesize}
\begin{small}
\begin{centering}
% \resizebox{\textwidth}{!}{
\begin{tabular}{lcll}
  \hline 
\bf  ID & \textbf{Feature} & \textbf{Type} & \textbf{Definition} \\ \hline
 &  & & \parbox[l]{6cm}{} \\
K & MEANC & \parbox[r]{1.5cm}{Numerical (100-700)} & \parbox[l]{6cm}{Meaningfulness rating of the word as provided by the Colorado norms of Toglia and Battig \cite{toglia-1978}} \\
  &  & & \parbox[l]{6cm}{} \\
L & MEANP & \parbox[r]{1.5cm}{Numerical (100-700)} & \parbox[l]{6cm}{Meaningfulness rating of the word as provided by the norms of Paivio (unpublished)} \\
 &  & & \parbox[l]{6cm}{} \\
M & AOA & \parbox[r]{1.5cm}{Numerical (100-700)} & \parbox[l]{6cm}{Age of acquisition, as provided by the norms of Gilhooly and Logie \cite{gilhooly-1980}.} \\
 &  & & \parbox[l]{6cm}{} \\
N & TQ2$_Q$ & Binary & \parbox[l]{6cm}{Morphological variant of another word in the dictionary.} \\
 &  & & \parbox[l]{6cm}{} \\
O & TQ2$_2$ & Binary & \parbox[l]{6cm}{Ends in the letter R and this R is not pronounced except when the next word begins with a vowel.} \\
 &  & & \parbox[l]{6cm}{} \\
P & WTYPE & Binary & \parbox[l]{6cm}{9 features indicating the word type (adverb, conjunction, interjection, adjective, noun, past participle, pronoun, verb, or other) as listed in the Shorter Oxford English Dictionary or Webster’s New International Dictionary.} \\
 &  & & \parbox[l]{6cm}{} \\
Q & STATUS & Binary & \parbox[l]{6cm}{7 features indicating the word status (archaic, alien, obsolete, colloquial, rare, and standard) as listed in the Dolby database \cite{dolby-1963}.} \\
 &  & & \parbox[l]{6cm}{} \\
R & STRESS & Binary & \parbox[l]{6cm}{14 features indicating the stress pattern of the word when pronounced. Where 2 is a strongly stressed syllable, 1 is medium stressed, and 0 is an unstressed syllable, the 14 stress patterns are: 0, 01020, 010200, 02, 020, 0200, 10020, 102, 1020, 10200, 20, 200, 2000, and 22.} \\
 &  & & \parbox[l]{6cm}{} \\
S & INFOBOX & Binary & \parbox[l]{6cm}{13 features indicating the type of infobox present in the English Wikipedia page for the word. Infobox types are: AMBIGUOUS, BIOGRAPHY\_VCARD, BIOTA, BORDERED, COLLAPSIBLE\_ AUTOCOLLAPSE, DEFAULT, GEOGRAPHY\_VCARD, HPRODUCT, NONE, VCARD, VCARD\_PLAINLIST, VEVENT, and VEVENT\_HAUDIO.} \\
 &  & & \parbox[l]{6cm}{} \\
T & Word Embeddings & Numerical & \parbox[l]{6cm}{300 features are the vector representation of the word derived using GloVe \cite{pennington2014glove}.} \\ 
% \hline 
\hline
\end{tabular} %}
\end{centering}
\end{small}
\caption{Features (K-T) used to represent words.} \label{table:wordFeaturesKT}
\end{table}

Features P and S (Table \ref{table:wordFeaturesKT}) can be categorised as high coverage (holding for more than two thirds of the tokens in the annotated corpora); features G, E, J, H, Q, and I (Tables \ref{table:wordFeaturesAJ} and \ref{table:wordFeaturesKT}) as medium coverage (holding for more than one third but less than two thirds of the tokens in the corpora); and features F, R, N, O, K, B, D, M, L, and O (Tables \ref{table:wordFeaturesAJ} and \ref{table:wordFeaturesKT}) as low coverage (holding for less than one third of the tokens in the corpora).\footnote{These features are listed in decreasing order of coverage provided.}

Considered individually, the great majority of features/feature sets listed in Table \ref{table:wordFeaturesAJ} have no linear relationship with the averaged human judgement of word complexity in the CWI 2016 and CWI 2018 datasets. The only exceptions are word length (feature group C) and the word's frequency count in the London-Lund corpus (feature group H). As the distributions of these two features are non-normal, we measured correlation with the averaged complexity ratings of words using Spearman's rho. We found that normalised word length has a low positive correlation ($\rho (28\,677) = 0.435, p < 0.001$) while the frequency of the word in the Brown corpus has a low negative correlation with word complexity ($\rho (28\,677) = -0.354, p < 0.001$). It is worth noting that MWEs in the CWI-2018 data are always complex and this may have influenced the results for word-length as MWEs are typically longer than single words.

%\todo[inline,backgroundcolor=yellow,size=\small]{Above, APA says not to insert 0 if the figure cannot be higher than 1. To me, the negative -.354 looks strange, so I added the 0s back in. - RJE} 

%

There is no linear relationship between the values of features/feature sets listed in rows K-S of Table \ref{table:wordFeaturesKT} and the averaged values of word complexity assigned by the annotators. In our experiments, we did not investigate the strength of correlations between individual word embedding features and average complexity ratings.

Given that the distributions of our features are non-normal, we used Levene's test \cite{levene-1960} to assess the homogeneity of variance between word feature values and complexity scores. In all cases, the Levene test statistic exceeded critical values and obtained $p<0.01$, indicating no equality of variance between complexity scores and feature values.

Clearly, this is a surprising result. Research in psycholinguistics indicates, for example, that the frequency of a given word (feature groups A, H, I, and J) affects its perception \cite{segui-1982,dupoux-1990,marslen-wilson-1990}, that word familiarity (feature group E) and frequency affect visual and auditory word recognition \cite{connine-1990}, and that word imageability (feature group G) significantly impacts word reading accuracy and rate of word learning among first and second graders at risk for reading disabilities \cite{steacy-2019}. Further, the word ``concreteness effect'' (feature group F) is a well-established concept in psycholinguistics with the tendency of words with tangible physical referents being learned earlier, recognised faster, and recalled with less effort than words with abstract referents \cite{paivio-1991,schwanenflugel-1991}. Schwanenflugel et al. \cite{schwanenflugel-1988} proposed that abstract words are more difficult to recognise because their interpretation is more reliant on context than is the case for concrete words. Word meaningfulness (feature groups K and L) has been observed to have a positive effect on word recognition \cite{leeds-1976} and words with great meaningfulness have been found to be easier to recall than words with less meaningfulness \cite{kinoshita-1989}. Finally, the age of acquisition of words (feature group M) has been reported to be a predictor of the speed of reading words aloud and lexical decision tasks (in which participants are asked to judge whether particular sequences of characters are real words), with words acquired early in life being responded to more quickly than words acquired later in life \cite{morrison-2000}. We would therefore expect to see more of our features correlating with complexity. This is likely to be a factor of the annotation protocols used in the datasets we analysed and motivates our wider argument in this work that there is a need for new CWI datasets. The two features that we did identify as showing correlation with word complexity (length and frequency) are both features that are used in almost all of the systems for the shared tasks at CWI--2016 and CWI--2018 as shown in Tables \ref{tab:approaches}  and \ref{tab:approaches2018} respectively. This indicates that these features are useful for complexity both in our correlation analysis and in the empirical results of the systems that have submitted using these features. We include this here to show the lack of correlation between sensible features and those datasets. In our next section, we will discuss the deficiencies of these datasets, as well as proposed our specification for an improved CWI dataset.

 \section{Specification for CWI Data Protocol}\label{sec:spec_cwi}

 In the previous Section we analysed differing features of complexity. In this section, we first highlight some of the design decisions that were taken in the creation of prior CWI datasets. We continue by proposing a specification, based on our prior analysis, for a new CWI dataset that improves on prior work. Our specification is designed to enable CWI research in areas that have not previously been explored. As well as providing a specification, we also provide a list of features for future datasets to implement in Table \ref{tab:cwi_spec}.
 
 \subsection{Building on Previous Datasets}\label{sec:building_on}
 
 The previous datasets for CWI have interesting characteristics that make them useful for the CWI task. A quick overview of these datasets is presented in Table \ref{tab:CWI_dataset_comparison}, where they are compared according to some of their basic features.
 
  \begin{table}[ht]
     \centering
     \begin{tabular}{cccccc}
     \hline
         Dataset & Binary & Probabilistic & Continuous & Context &  Multi-Genre\\\hline
         CWI--2016 & $\times$ & & & $\times$ & $\times$\\
         CWI--2018 & $\times$ & $\times$ & & $\times$ & $\times$\\
         Maddela--2018\cite{maddela-xu-2018-word} & & & $\times$ & &\\
         \hline
     \end{tabular}
     \caption{CWI Datasets compared according to their features. `Binary', `Probabilistic' and `Continuous' refer to the nature of the annotated labels. `Context' refers to the presence of sentential context at annotation time and `Multi-Genre' refers to the dataset drawing from sources across many genres.}
     \label{tab:CWI_dataset_comparison}
 \end{table}
 
 %2016 annotation protocol
 The first dataset we have considered is the CWI--2016 dataset, which provides binary annotations on words in context. 9,200 sentences were selected and the annotation was performed as described below in \cite{paetzold-specia:2016:SemEval1}:
 
 \begin{quote}
 \textit{
 ``Volunteers were instructed to annotate all words  that  they  could  not  understand\dots A subset of 200 sentences was split into 20 sub-sets of 10 sentences, and each subset was annotated by a total of 20 volunteers. The remaining 9,000 sentences were split into 300 subsets of 30 sentences, each of which was annotated by a single volunteer.''
}
 \end{quote}
 
 The annotators were asked to identify any words for which they did not know the meaning. Each annotator had a different proficiency level and therefore will find different words more or less complex - giving rise to a varied dataset with different portions of the data reflecting differing complexity levels. Further, each instance in the test set was annotated by 20 annotators, whereas each instance in the training set was annotated by a single annotator. For the test set, any word which was annotated as complex by at least one annotator was marked complex (even if the other 19 annotators disagreed). This is problematic as the training data is not representative of the testing data, making it hard for supervised systems to do well on this task. 
% Binary annotation of complexity appears to involve a mental thresholding task based on the annotator's judgment of complexity. 
 Binary annotation of complexity requires an annotator to impose a subjective threshold on the level at which they transition from considering a word complex as opposed to simple.
 An annotator's background, education, etc. may affect where this threshold between complex and simple terms should be set. Further, it is likely that one annotator may find words difficult that another finds simple and vice-versa. Factors such as the annotator's native language, educational background, dialect, etc. all affect the type of words they are familiar with. In the case of the training data where 20 annotators have all annotated the same instance and any instance with at least one annotation is considered complex, it may be taken that the annotations represent some form of maximum complexity - i.e., that any word is above the lowest possible threshold of complexity. However, in the case of the test set where each word is annotated by a single annotator, the annotations are harder to interpret. Each instance is personal, reflecting only a single annotator's judgment.
 
 %2018 ST annotation protocol
 Moving on from the CWI--2016 dataset, the CWI--2018 dataset also provides binary annotations, which were aggregated to give a `probabilistic' measure of complexity. CWI--2018 invited participants to submit results on both the binary complexity annotation setting and the probabilistic annotation setting. To collect their data, the organisers of CWI--2018 followed a similar principle as in CWI--2016. Sentences were presented to annotators and the annotators were asked to select any words or phrases that they found to be complex. As in CWI--2016, the annotation task in CWI--2018 was subjective, with potentially low agreement between annotators. In the probabilistic setting, at least 20 annotations were collected from native and non-native speakers and each word was given a score indicating the proportion of annotators that found that word to be complex. (i.e., if 10 out of 20 annotators marked the word, then it would be given a score of 0.5). A useful property of this style of annotation is that words are seen on a probablistic scale of complexity. However, the aggregation of binary annotations to give continuous annotations does not necessarily tell us  about the complexity of the word itself. Instead it tells us about the annotators, and how many of them will consider a word complex. So, for example a score of 0.5 does not indicate a median level of complexity (or some sort of neutrality between simple and complex), but instead should be interpreted as indicating that 50\% of the annotator pool will consider this word complex.
 
 %2018 Maddela annotation protocol
 
 The final dataset we have covered was published in 2018 by Maddela and Xu \cite{maddela-xu-2018-word}. We refer to this as Maddela--2018 for brevity.
 In this dataset, 11 annotators who spoke English as a second language were employed to annotate a portion of 15,000 words on a 6-point Likert scale with 5--7 annotations being collected for each vocabulary item. 
 Words were presented without context, with the annotators guessing or making assumptions about the sense of the word at annotation. Different annotators may have considered the word to have a different sense or to have been used in a different context. 
 Almost all words are polysemous and the different senses of the words are likely to have different levels of complexity - particularly in a coarse grained sense setting (e.g., \textit{mean} average vs. a \textit{mean} person). 
 The main effect here is that the varied complexities of the multiple senses and usages of a word are conflated into a single annotation. 
 There is no information as to which word sense the annotators were giving the annotations for, and as such the annotations may be unreliable in cases where a word is used in an uncommon sense. 
 In the Likert-scale type annotation, it is less of an issue that annotators' opinions will vary than in the binary setting used in CWI--2016 and CWI--2018, as each annotator's judgment is aggregated on a common continuous scale. 
 This means that the final averaged annotation is reflective of the average complexity that a word might have in a general setting. 
 This is making an assumption that the annotations are normally distributed and that a mean average is valid in this case. A normality test could be used to quantify whether instances are likely to have normal distributions, however with only 5--7 annotations per instance, this may not be reliable.
 
 % MWE treatment
 
 So far, we have mainly considered complex words. However, the complexity of multi-word expressions is a valuable addition to the CWI literature. MWEs can be considered as compositional or non-compositional. Compositional MWEs (e.g., christmas tree, notice board, golf cart, etc.) take their meaning from the constituent words in the MWE, whereas non-compositional MWEs do not (e.g., hot dog, red herring, reverse ferret,etc.). It is reasonable to assume that complexity will follow a similar pattern to semantics and that compositional MWEs will be dependent on the constituent words to give the complexity of the expression, whereas the complexity of non-compositional MWEs will be independent of the constituent words. In the previous datasets, only the CWI--2018 dataset asked annotators to highlight phrases as well as single words, giving a limit of 50 characters to prevent overreaching. Participants in the task were asked to also give complexity annotations for the highlighted phrases. The system with the highest overall score reported that they found it easier to always consider MWEs as complex in the binary setting \cite{syspaper7}. The work of \cite{maddela-xu-2018-word} also considers MWEs. Although they do not annotate for these, instead using average pooling to combine the embeddings of each token in a phrase into a single embedding, which is then processed in the same way as for single words. As described previously, this assumes compositionality, which will not always be the case.
 
 % POS treatment
 Little treatment has been given to the variations in complexity between different parts of speech. None of the previous datasets annotate specifically for part of speech except for the CWI--2016 shared task data, which explicitly asks annotators to only highlight content words in the target sentences. Again, this is an important consideration as the roles of nouns, verbs, adjectives and adverbs are different in a sentence and considering them as different entities during annotation will help to better structure corpora. Developers of the existing corpora that span POS tags all suggest the use of POS as a feature for classification --- demonstrating its importance in CWI.
 
 % Native vs. non-native speakers
 All of the corpora recognise the importance of a diversity of reader backgrounds in their corpus construction. Native speakers of English might not realise that certain words they know well (depending on their socio-cultural biases) are not commonly known or may falsely assume that they ``find all words easy''. All three of the corpora that we have studied include annotations by non-native speakers. The CWI--2016 dataset used crowdsourcing to get annotations from 400 non-native speakers, the CWI--2018 dataset used native and non-native speakers (collecting at least 10 annotations from each for every instance). The Maddela--2018 data used 11 non-native speakers. The use of non-native speakers for CWI annotation may lead to models trained using these datasets being useful for identifying words which are complex to non--native speakers, but may not be applicable to other groups.

 % Text genres covered
 All the datasets are heavily biased towards text which has not been professionally edited. The CWI--2016 dataset compiles a number of sources taken from Wikipedia and Simple Wikipedia, the CWI--2018 dataset takes Wikipedia, WikiNews and one formal set of news text sources. The Maddela-2018 dataset uses the Google Web1T \cite{brants2006google} (taken from a large web-crawl) to identify the most frequent 15,000 words in English and re-annotates each for complexity. Except for the news texts in the 2018 data, all of these sources are written for informal purposes and will contain spelling mistakes, idioms, etc. There has been little prior work exploring cross-genre learning for CWI, however it is unlikely that models trained on such informal text will be appropriate for identifying complexity in formal texts.
 
 \subsection{Specification}\label{sec:specification}
 
In the remainder of this section we will describe some of the qualities of an ideal dataset for CWI. Our recommendations are summarised at the end of this Section in Table \ref{tab:cwi_spec}. This specification is intended to give general purpose recommendations for anyone seeking to develop a new CWI dataset. 
 
 % continuous
 The key issue with the shared task datasets was the subjectivity that arose during the annotation process due to their treatment of complexity as a binary notion. When multiple annotators are asked to ``mark any complex word'' they will each draw on their subjective definition of complexity, and each will choose a different subset of words to be annotated as complex.
 The annotations that result from this are probabilistic in nature and tell us more about the annotators than the words themselves. Future datasets should consider providing measures which attempt to give more objectivity and move towards consensus between annotators. Of course, any complexity annotation involving human participants will always rely on the participants subjective knowledge and hence will be dependent on the participants.
 More objective measures of continuous complexity could be given by asking annotators to mark words on a Likert scale as by Maddela-2018, or by looking at external measurements of the ability of people to read the words, such as lexical access time, eye tracking, etc. 
 
 There are two factors to be considered here when measuring word complexity. One is the perceived complexity of a word (how difficult an annotator estimates a word to be) and the other is the actual complexity of a word (how much difficulty that word presents to the reader) \cite{leroy2013user}. Clearly these are both important factors in estimating a word's complexity and although we may expect them to be correlated there is no guarantee they will be aligned. Whereas perceived complexity affects how a user may prejudge a text, actual complexity determines the degree with which a reader is likely to struggle.
 
 Of course, any measure of complexity which is derived by asking humans to give a subjective judgment of how difficult they find a word is bound to give a measure of perceived rather than actual complexity. In fact, measuring actual complexity would only be possible if the human was taken out of the loop altogether (even a setting where the reader doesn't know they are being assessed would rely on a participant's innately subjective assessment of each word). Any annotation scheme which focusses on continuous complexity judgments is still inviting perceived complexity assessment. By giving more levels to the assessment of complexity (i.e., through a Likert Scale assessment) the annotators have more ability to better record their perception of the complexity of the words that are being assessed.

 % context
 
 The only previous dataset to present continuous annotations (Maddela-2018) did so in the absence of context. Context is key to determining the usage and meaning of a word and the same word used in different contexts can vary greatly in both semantics and complexity. Indeed, a familiar word in an unfamiliar context may be just as jarring as a rare word for a reader, who is forced to quickly update their mental lexicon with the new sense of the word they have encountered (e.g. words like \emph{base}, \emph{boss}, and \emph{fanning} in the domains of chemistry, architecture, and geology, respectively). Datasets should include context for any words that annotations are provided for. This will help systems to identify how contextual factors affect the complexity of a given instance. When presenting context, researchers may wish to either ask annotators to mark every word in a sentence according to some complexity judgment (dense annotation) or they may wish to pick a target word in a context and ask only for a judgment of the complexity of this word (sparse annotation). In the dense annotation setting, it is likely to be possible to get a much higher throughput of complexity annotations, as the reader will need to only read a sentence once to give multiple annotations, however they are likely to be deeply influenced by the meaning of the sentence, and may struggle to disassociate this from their annotation of complex words themselves. In the sparse annotation setting more contexts are required to give a comparative number of instances compared to the dense annotation setting, however the annotation given is more likely to be a direct result of the token itself, rather than the sentence. Any such sparse annotation task should be set up to ensure that an annotator gives judgments based on the word in its context (i.e., that they read and understand the context), rather than just giving a judgment based on the word, as if no context were presented. 
 
 % multiple instances of each word

 Given that we are recommending that the data is presented in context, there is a strong argument for presenting multiple instances of each word. If only one instance of a word were presented in context, then it may be the case that this word had a specific usage that was not representative of its general usage. Words are polysemous \cite{Fellbaum2010} and this is true both at the coarse grained  (tennis \textit{bat} vs. fruit \textit{bat}) and narrow grained levels (I \textit{love} you vs. I \textit{love} London). The coarse grained level represents different meanings or etymologies, whereas the fine-grained level may represent a similar meaning but a different intensity (as in our example). The provision of multiple instances of a word allows both of these factors to be taken into account. This consideration should be held in balance with the need to have a diversity of tokens. If a dataset has $N$ instances, constituting $P$ occurrences of $R$ words, then we suggest that $R \gg P$. I.e., the number of total words should be much larger than the number of instances of each word. There is more to be gained in a dataset by having a diversity of tokens than by having many annotations on each token. An interesting separate task would be to annotate many instances of one word form for complexity and analyse how the context affects this. However, this is a secondary task to the one we are presenting here of assessing a word's complexity.

 % multiple annotations (as many as possible)
 
 Each instance in a new CWI dataset should be viewed and annotated by multiple people, ideally from a spectrum of ability levels. Multiple annotations have been a common theme of the previous CWI datasets we  have discussed, with datasets using as many as 20 annotators per instance. All subsets (train, dev, test) of a dataset should be annotated by the same number of annotators, or at the very least by annotators drawn from the same distribution. This ensures that all subsets of the data are comparable. 
 More annotations allows us to capture a wider array of viewpoints from annotators of varying ability levels. If the annotators are carefully selected to ensure they represent a mixture of ability levels then this will lead to annotations that are representative. Consider the case where all annotators are of low ability, or of high ability. The resulting annotations may lead to all words being assigned to the most or least complex categories respectively. This may be desirable in user- or genre-specific settings, but is not desirable for general-purpose LCP. There are two potential approaches to selecting a pool of annotators and distributing annotations between them. Firstly, a researcher may choose to use a fixed number of annotators, such that each annotator views every data instance once. In this setting, each data instance receives N annotations, where N is the number of annotators chosen. Secondly, the annotations may be distributed across a wider pool of annotators, where given N annotators each sees a randomised subset of the data. In this setting, a researcher may choose to control how many instances each annotator sees, ensuring an even distribution of annotators across the data instances. The second approach is more appropriate in a crowd-sourcing setting, where a researcher has diminished ability to control who takes on which job.
 
 % Mixture of native and non-native speakers
 
 %\todo[]{We might also keep in mind that both native and non-native speakers may simultaneously be specialists in some domains and novices in other domains. Non-natives may be specialists in domains where natives are not, and vice versa.}
 Previous CWI datasets for English have given a strong focus on non-native speakers as discussed above. Non-native speakers have learnt English as a foreign language and the assumption in using them for CWI research is that they will have only learnt a simple subset of English that allows them to get by in daily tasks. However, a non-native speaker may range from a new immigrant who has recently arrived in an English speaking country to someone who has lived there for decades. Further, both native and non-native speakers may simultaneously be specialists in some domains and novices in other domains. Non-natives may be specialists in domains where natives are not, and vice versa, influencing their complexity judgments. We would suggest, that whilst non-native speakers should not be excluded from the CWI annotation process, they should not be relied upon either. Instead the pool of annotators should be selected for their general ability in English, not for their mother tongue. Indeed, when selecting non-native speakers it may be worth considering selecting a variety of mother-tongues, as it is the case that different languages, or language families will have cognates and near-cognates with English, making it easier for non-native speakers of certain backgrounds to understand words in English with roots in their  mother tongue.
 
 % multi-genre (including professionally written text)
 Allowing for multiple genres gives more diversity in the type of text studied and allows systems that are trained on it to generalise better to unseen texts. This prevents overfitting to one text-type, leading to results being more reliable and hence more interpretable, and ultimately leads to the creation of useful models that can be applied across genres. CWI resources should name the source genres that their texts are taken from and comply with licences placed on those genres. Whilst informal, or amateur text is in abundance (e.g., Twitter or Wikipedia), formal texts should also be considered for CWI such as professionally written news, scientific articles, parliament proceedings, legal texts or any other such texts that are written for a professional audience. These texts provide well structured language, which is typically targeted at a specific audience  and is of a difficult quality for those outside that audience. These texts contain a higher density of complex words and as such are useful examples of the types of text that might need interventions to improve their readability for a lay reader.

 % Treatment of MWEs
 As discussed previously, MWEs are an important element in complexity as previous studies have shown that MWEs are generally considered more complex by a user than individual words \cite{syspaper7}. Any new CWI dataset should consider incorporating MWEs as they will certainly be useful for future CWI research. When we consider that MWEs can range from simple collocations (White House), to verbal phrases (pick up) and may span 2 or more words, across parts of speech --- including phrasal MWEs (it's raining cats and dogs)  --- it is clear that the number of potential MWEs to consider is much wider than the number of single tokens. How do we select appropriate MWEs to cover? There is no particular advantage to CWI in selecting one category of MWE over another, but we suggest that any dataset covering MWEs explicitly names the types of MWE that it has covered. By incorporating MWEs, a dataset may be used to investigate both the nature of complexity in those MWEs and in the constituent tokens. Strategies for identifying MWEs, as well as the different types of MWEs are beyond the scope of this work and we would direct the reader to the MWE literature \cite{sag02,schneider-etal-2014-comprehensive} for a more comprehensive treatment of this problem.

% style this as a table instead of a list? 
\begin{table}[ht]
    \centering
    \begin{tabular}{ccp{6cm}}
        \hline
        \bf ID & \bf Feature & \bf Description \\\hline
        1 & Continuous annotations & Complexity labels should be on a continuous scale ranging from least to most difficult. \\
        2 & Context & Tokens should be presented in their original contexts of usage.\\ 
        3 & Multiple token instances & Each token should be included several times in a dataset.\\
        4 & Multiple token annotations & Each token should receive many annotations from different annotators\\
        5 & Diverse annotators & The fluency and background of annotators should be as diverse as possible.\\
        6 & Multiple genres & The text sources used  to select contexts should cover diverse genres.\\
        7 & Multi-word expressions & These should be considered alongside single word tokens as part of an annotation scheme.\\
        \hline
    \end{tabular}
    \caption{A list of recommended features for future CWI dataset development.}
    \label{tab:cwi_spec}
\end{table}
 % continuous
 % context
 % multiple instances of each word
 % multiple annotations (as many as possible)  % annotators of varied backgrounds
 % Mixture of native and non-native speakers
 % multi-genre (including professionally written text)
 % Treatment  of MWEs

\section{CompLex 2.0}\label{sec:complex2}
%intro

In this Section we describe a new CWI dataset that we have collected. Our new dataset, dubbed `CompLex 2.0' builds on prior work (CompLex 1.0 \cite{shardlow-etal-2020-complex}), in which we collected and annotated tokens in context for complexity. We have described the data collection process for CompLex 1.0 as below and then the annotation process that we undertook to extend this data to CompLex 2.0. CompLex 2.0 covers more instances than CompLex 1.0 and crucially, has more annotations per instance than CompLex 1.0, making it more reliable. We present statistics on  our new dataset and describe how it fits the recommendations we have made in our specification for new CWI datasets above. CompLex 2.0  was used as the dataset for the SemEval Shared Task on Lexical Complexity Prediction in 2021.

\subsection{Data Collection}

The first challenge in dataset creation is the collection of appropriate source texts. We have followed our specification above and selected three sources that give a sufficient level of complexity. We aimed to select sources that were sufficiently different from one another to prevent trained models generalising to any one source text. The sources that we used are described below. 

\begin{itemize}
 \item \textbf{Bible:} We selected the World English Bible translation \cite{Christodouloupoulos2015}. This is a modern translation, so does not contain archaic words (thee, thou, etc.), but still contains religious language that may be complex. The inclusion of this text  gives language that combines narrative and poetic text that uses language typically familiar for a reader, yet interspersed with unfamiliar named entities and terms with specific religious meanings (\emph{propitiation}, \emph{atonement}, etc.).
 
 \item \textbf{Europarl:} We used the English portion of the European Pariliament proceedings selected from europarl \cite{koehn2005europarl}. This is a very varied corpus concerning a wide range of issues related to European policy. As this is speech transcription, it is often dialogical in nature in contrast to our other two corpora. Again, the style of text is generally familiar as it is transcriptions of debates. However technical terminology relating to the topics of discussion is present, raising the difficulty level of this text for a reader.
 
 \item \textbf{Biomedical:} We selected articles from the CRAFT corpus \cite{bada2012concept}, which are all in the biomedical domain. These present a very specialised type of language that will be unfamiliar to non-domain experts. Academic articles present a classic challenge in understanding for a reader and  are typically written for a very narrow audience.  We expect these texts to be particularly dense with complex words. 
\end{itemize}

In addition to single words, we also selected targets containing two tokens. We used syntactic patterns to identify these MWEs, selecting for adjective-noun or noun-noun patterns. We discounted any syntactic pattern that was followed by a further noun to avoid splitting complex noun phrases (e.g., noun-noun-noun, or adjective-noun-noun). We used the StanfordCoreNLP tagger \cite{manning2014stanford} to get part-of-speech tags for each sentence and then applied our syntactic patterns to identify candidate MWEs.

Clearly this approach does not capture the full variation of MWEs. It limits the length of each to 2 tokens and only identifies compound or described nouns. Some examples of the types of MWE that we identify with this scheme are given in Table \ref{tab:mwes}. Whilst this inhibits the scope of MWEs  that are present in our corpus, this does allow us to make a focused investigation on these types of MWEs. Notably, the types of MWE  that we have identified are those that are the most common  (compound nouns, described nouns, compositional,  non-compositional and named entities). The investigation of other types of MWEs may be addressed by other, more targeted studies following our recommendations for CWI annotation.

\begin{table}[ht]
    \centering
        \begin{tabular}{ccc}
        \hline
           \bf Pattern & \bf MWE & \bf Type  \\\hline
            NN & storage box & Compound Noun \\
            JN & ready meal & Described Noun \\
            JN & electric vehicle & Compositional\\
            NN & hot dog & Non-compositional\\
            JN & European Union & Named Entity\\
            \hline
        \end{tabular}
    \caption{The varied types of MWEs that can be captured by our syntactic pattern matching.  NN  indicates a Noun-Noun pattern, whereas JN indicates an Adjective-Noun pattern.}
    \label{tab:mwes}
\end{table}

%We have not analysed the distribution of compositional vs. non-compositional constructions in our dataset, however we expect both to be present. It would be interesting to further analyse these to distinguish whether the complexity of an MWE can be inferred from tokens in the compositional case, and to what degree this assumption holds for the non-compositional case.

%int[] lowerBounds = { 2, 5, 11, 51, 251, 501, 1401, 3101 };
%int[] upperBounds = { 4, 10, 50, 250, 500, 1400, 3100, 10000 };

For each corpus we selected words using frequency bands, ensuring that words in our corpus were distributed across the range of low to high frequency. 
We selected the following eight frequency bands according to the SUBTLEX  frequencies in order of least to most frequent (i.e., most to least complex): 2--4, 5--10, 11--50, 51--250, 251--500, 501--1400, 1401--3100, 3101--10000. 
We excluded the rarest words (those with a frequency of only 1) as well as the most frequent (those above  10,000) in order to ensure that our instances were well-attested content words.
As frequency is correlated to complexity \cite{brysbaert2011word}, this ensures that our final corpus will have a range of high and low complexity targets. We chose to select 3000 single words and 600 MWEs from each corpus to give a total of 10,800 instances in our corpus. We selected a  representative number of instances from each frequency band to give the desired total number of instances in each corpus. We automatically annotated each sentence with POS tags and only selected nouns as our targets, in-keeping with our MWE selection strategy. We allowed a maximum of 5 instances of a token to be selected in each genre (ensuring that contexts were different). This maximises the total number of examples of each instance, whilst still allowing some variation  in the selection of tokens. There is a theoretical minimum of 600 instances  of single words and 120 MWEs that could occur in our corpus (each with 5 occurrences in each of the three genres. Table \ref{tab:complex2.0} shows that the number of repeated instances is much lower. This is a factor of the stochastic selection procedure that we have employed. We have included examples of the contexts and target words in Table \ref{tab:examples}.

\begin{table*}[ht]
    \centering
  
    \begin{tabular}{lp{8cm}c}
    \hline
        \multicolumn{1}{c}{\textbf{Corpus}} & \multicolumn{1}{c}{\textbf{Context}} & \textbf{Complexity}  \\ \hline
        Bible & This was the \textbf{length} of Sarah's life. & Low \\
         Biomed & [...] cell \textbf{growth} rates were reported to be 50\% lower [...] & Low \\ 
        Europarl & Could you tell me under which rule they were enabled to extend this item to have four rather than three \textbf{debates}? & Low \\
        Europarl & These agencies have gradually become very important in the \textbf{financial world}, for a variety of reasons.	 & Medium\\
        Biomed & [...] leads to the \textbf{hallmark loss} of striatal neurons [...] & Medium \\
        Bible & The \textbf{idols} of Egypt will tremble at his presence [...] & Medium \\
           Bible & This is the law of the \textbf{trespass offering}. & High \\
          Europarl & They do hold elections, but candidates have to be endorsed by the conservative clergy, so \textbf{dissenters} are by definition excluded.& High \\
        Biomed & [..] due to a reduction in \textbf{adipose} tissue. & High \\
    \hline
    \end{tabular}
    
    \caption{Examples from our corpus, the target word is highlighted in bold text. The field {\em Complexity} refers to perceived complexity}
    \label{tab:examples}
\end{table*}

\subsection{Data Labelling}

%  - data annotation
As has been previously mentioned, prior datasets have focused on either (a) binary complexity or (b) probabilistic complexity. Neither of which give a true representation of the complexity of a word.  In our annotation we chose to annotate each word on  a 5-point Likert scale, where each point was given the following descriptor:
\begin{description}
 \item[1. Very Easy:]  Words which were very familiar to an annotator.
 \item[2. Easy:] Words for which an annotator was aware of the meaning.
 \item[3. Neutral: ]  A word which was neither difficult nor easy.
 \item[4. Difficult:] Words for which an annotator was unclear of the meaning, but may have been able to infer the meaning from the sentence.
 \item[5. Very Difficult:] Words that an annotator had never seen before, or were very unclear.
 
\end{description}

We used the following key to transform the numerical labels to a 0-1 range when aggregating the annotations: $1 \rightarrow 0$, $2 \rightarrow 0.25$, $3 \rightarrow 0.5$, $4 \rightarrow 0.75$, $5 \rightarrow 1$. This allowed us to ensure that our complexity labels were normalised in the range 0--1.

%   - crowd sourcing

We initially employed crowd workers through the Figure Eight platform (formerly CrowdFlower), requesting 20 annotations per data instance and paying \$0.03 per annotation. We selected annotators from English speaking countries (UK, USA and Australia). In addition, we used the annotation platform's in-built quality control metrics to filter out annotators who failed pre-set test questions, or who answered a set of questions too quickly. 

%Our job completed within 3 hours, with over 1500 annotators. The annotators were able to fill in a post-hoc annotation survey, with average satisfaction being around 3 out of 5, the scores typically lower on the `ease of job' metric.

%   - filtering post crowd-sourcing

After we had collected these results, we further analysed the data to detect instances where annotators had not fully participated in the task. We specifically analysed instances where an annotator had given the exact same annotation for all instances (usually these were all 'Neutral') and discarded these from our data. We retained any data instance that had at least 4 valid annotations in our final dataset.

This led to the version of the dataset we described as CompLex 1.0. Whilst this dataset evidenced the trends we expected to see, the conclusions we were able to draw from it were weaker than we hoped  \cite{shardlow-etal-2020-complex}. The median number of annotators was 7 per instance, and we identified this as an area for improvement. The involvement of more annotators would allow more opinions to be expressed, leading to better average judgments.

For the second round of annotations we used the Amazon Mechanical Turk platform. We used exactly the same data as in the original annotation of CompLex 1.0 and requested new annotations for each instance. We gave the  same  instructions to  annotators regarding the Likert-scale points. As there is no in-built quality  control in Mechanical Turk, we opted to release the  data in  batches  (1200 instances  at a time). We asked for a further 10 annotations per instance and paid at a rate of  \$0.03 per annotation. We reviewed the annotators work  in between batches, rejecting accounts which submitted annotations too quickly,  or without correlation to the other annotator's judgments. We also measured the correlation with lexical frequency to ensure that the annotations we were receiving  were in the  range we expected. 

This allowed us to gather a further 108,000 annotations on the CompLex data. These  new judgments were aggregated with those from CompLex 1.0 to give a new dataset --- CompLex 2.0. We used this data to run a shared task on Lexical Complexity Prediction at SemEval 2021 \cite{shardlow2021semeval}.
%\todo[]{Can we cite anything yet, even if it is ``forthcoming'' or perhaps the URL?}. 

\subsection{Corpus Statistics}

%Stats on Complex 1.0
The first round  of annotations led to an initial version of the Corpus (CompLex 1.0), for which we have shown the statistics originally reported in Table \ref{tab:complex1.0}. Due to the quality control that we employed for this round of annotation, we discarded a large portion of our original judgments and only kept instances with four or more annotations. This is evident in the fact that only 9,476 instances out of our original 10,800 are present in this iteration of the corpus.  Additionally, the median number of annotators was 7 across our corpus (with the range being from 4 to 20). Retaining only the annotations in which we could be certain of the quality was a difficult choice, as it reduced the amount of data available.  However, the mean complexities of the sub-corpora were in  line with our expectations. With Biomedical text being on average more complex than the other two genres. 

\begin{table}[ht]
    \centering
    \begin{tabular}{lccc}
        \hline
        \bf Genre & \bf Contexts & \bf Unique Words  & \bf Complexity \\\hline
        All      & 9,476 & 5,166  & 0.394 \\
        Europarl & 3,496 & 2,194  & 0.390 \\
        Biomed   & 2,960 & 1,670  & 0.407 \\
        Bible    & 3,020 & 1,705  & 0.385 \\
        \hline
  
    \end{tabular}
    \caption{The statistics for CompLex  1.0. We report on the entire corpus and also present a breakdown of statistics by \textbf{Genre}. We include statistics on the number of \textbf{Contexts}, the number of \textbf{Unique Words} and the mean \textbf{Complexity} in each partition}
    \label{tab:complex1.0}
\end{table}

%Stats on AMT annotation
This led us to undertake our second round of annotation in order to develop CompLex 2.0 ready for the SemEval shared task. We have included statistics on the annotations aggregated from both rounds in Table \ref{tab:amt_stats}. 513 separate annotators viewed our data,  with each annotator seeing on average 542 instances across all rounds of annotation (around 5\% of our corpus). We gathered a total of 278,093 annotations, paying \$0.03 per annotation. The average time spent per annotation was  21.61 seconds, which means that we paid our workers at an average rate of 5 US Dollars per hour. The task received reviews indicating that annotators found it to be well paid in comparison to other tasks on the platform. We gathered an average of 25.75 annotations per instance,  this is an increase over CompLex 1.0, which only had on average 7 annotations per instance. We expect that by having more annotations per instance, we will have more reliable average estimates of the complexity of each word.

\begin{table}[ht]
    \centering
    \begin{tabular}{cc}
        \hline
        Number of Annotators & 513 \\
        Number of Instances & 10,800 \\
        Number of Annotations & 278,093 \\
        Annotations per Instance & 25.75 \\
        Instances per Annotator & 542.09\\
        Time per Annotation & 21.61 (s)\\
        \hline
    \end{tabular}
    \caption{Statistics on the round of evaluation undertaken with Mechanical Turk.}
    \label{tab:amt_stats}
\end{table}

%Stats on Complex 2.0

We report detailed statistics on our new dataset, CompLex 2.0, in Table \ref{tab:complex2.0}. We can see that in total  5,617 unique tokens covering single words and multi-word expressions are distributed across 10,800 contexts. Whilst the contexts are split evenly between each genre (3,600 each) the number of repeated words is higher in the Biomed and Bible corpora, with more distinct words occurring in the Europarl corpus. The complexity annotations are low at 0.321 for the entire corpus, indicating that the average complexity of words is somewhere between points 2 (0.25 --- a word which that the annotator was aware of the meaning) and 3 (0.5 --- A word which was neither difficult nor easy) on our Likert scale. This indicates that annotators generally understood the words in our dataset. The annotations did use the full range of our Likert scale and the dataset contains  words of all complexities. We can see from the data that the Biomedical genre  was on average more difficult to understand (0.353) than the other genres (0.303 for Europarl and 0.307 for Bible respectively). Multiword expressions are markedly more complex (0.419) than single words (0.302), with the same genre distinctions as in the full data.

\begin{table}[ht]
    \centering
    \begin{tabular}{llccc}
        \hline
        \bf Subset & \bf Genre & \bf Contexts & \bf Unique Words  & \bf Complexity \\\hline
        \multirow{4}{*}{All} &        
         \bf Total     &  \bf 10,800 &  \bf 5,617 &  \bf 0.321 \\ %\cline{2-5}
        & Europarl & 3,600  & 2,227 & 0.303 \\
        & Biomed   & 3,600  & 1,904 & 0.353 \\
        & Bible    & 3,600  & 1,934 & 0.307 \\\hline
        \multirow{4}{*}{Single} &        
         \bf  Total      &  \bf 9,000  &  \bf 4,129 &  \bf 0.302 \\ %\cline{2-5}
        & Europarl & 3,000  & 1,725 & 0.286 \\
        & Biomed   & 3,000  & 1,388 & 0.325 \\
        & Bible    & 3,000  & 1,462 & 0.293 \\\hline
        \multirow{4}{*}{MWE} &        
         \bf  Total     &  \bf 1,800 &  \bf 1,488  &  \bf 0.419 \\ %\cline{2-5}
        & Europarl & 600   & 502   & 0.388 \\
        & Biomed   & 600   & 516   & 0.491 \\
        & Bible    & 600   & 472   & 0.377 \\
        \hline
    \end{tabular}
    \caption{The statistics for CompLex 2.0. We report on the entire corpus and also present a breakdown of statistics by \textbf{Genre} and by \textbf{Subset}. We include statistics on the number of \textbf{Contexts}, the number of \textbf{Unique Words} and the mean \textbf{Complexity} in each partition}
    \label{tab:complex2.0}
\end{table}

\subsection{Inter-annotator Agreement}
%use a normality test to see whether annotations are typically normally distributed
% find examples where they are and where they are not.

Achieving strict adherence to annotation guidelines is difficult in the crowd-sourcing setting as there is little time to train, test or survey annotators. As a result, inter-annotator agreement tends to be lower in this context. We provided some controls as outlined above to ensure that annotators were fully participating in the task and that their annotations aligned with those of other annotators. In our setting, we do not necessarily expect annotators to agree in every case as one may legitimately consider a word to be complex, whilst another considers it to be simple. A reasonable expectation is that annotators will provide similar annotations to each other, and that the annotations will mostly fall into one category. We expect the distribution of annotations for one instance to be normally distributed. We have already made this assumption, as we take the mean to give the average complexity.

\begin{figure}[ht]
    \centering
    \includegraphics[width=\textwidth]{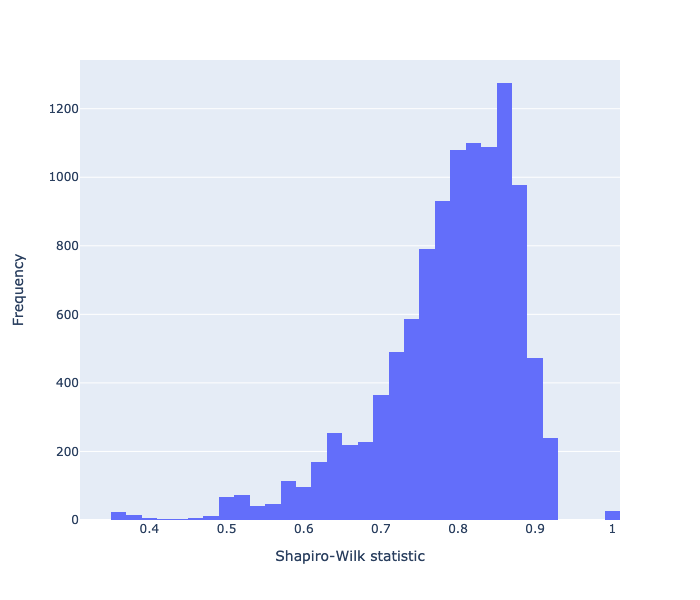}
    \caption{A histogram of Shapiro-Wilk's test statistics, demonstrating the likelihood for each instance that the annotations are normally distributed.}
    \label{fig:Shapiro_Histogram}
\end{figure}

To test this, we used a Shapiro-Wilk test \cite{shapiro_wilk}, which gives a number in the range of 0-1 indicating how likely a given distribution is to follow the normal distribution. For each of our instances, we perform the test on the annotations for that instance. A higher number indicates that the instance has annotations which are more likely to be normally distributed, whereas a low number on this test indicates a non-Gaussian distribution, such as a multi-modal distribution. A histogram of this data is displayed in Figure \ref{fig:Shapiro_Histogram}. This shows that the majority of our data obtains a score between 0.7 and 0.9 according to the Shapiro-Wilk test, with a peak around 0.85. This indicates that our data is generally normally distributed, and hence that annotators generally gave annotations that centered around a mean value.

In Table \ref{tab:Shapiro_examples} we have shown a number of examples  from our corpus that do not follow the distribution that we may have expected. These were infrequent in our corpus, but are displayed here to help the reader understand where annotators may have disagreed.  In example 1 the simple word `heaven' was given to annotators, most of whom assigned it to the \textit{Very Easy} category. However, 3 annotators disagreed with this, assigning it to the \textit{Neutral} category. Possibly, the annotators found the word easy, but the metaphorical usage harder to grasp. Example 2 shows a similar disagreement, albeit  around a more difficult word. `Election' is a word that most people living in a democracy will have encountered, yet 5 people felt it was neither easy, nor difficult --- placing it in the \textit{Neutral} category. Our third example, taken from the Biomedical genre, demonstrates a word (Granules) which is considered \textit{Easy} by 14 annotators, yet is considered \textit{Difficult} by 4 annotators. Whilst `Granules' is not a particularly rare word, it may be considered complex by some in this instance due to its contextual usage in the biomedical literature.  Example 4 shows a word which is specific to biblical language (`Cubit').  Although the annotations gave a reasonably Gaussian set of annotations  (0.848 according to the Shapiro-Wilk statistic), they were split over all 5 potential categories. This is an example of annotators' previous familiarity with the text. Those who know a cubit is an ancient measure of length will score it on the easier side of the Likert scale, whereas those who have not seen the word before will score it as more difficult. The remaining three examples (5--7) all score similarly highly on the Shapiro-Wilk test, however they have a wide spread of annotations. Again, this is likely due to the familiarity of the annotators with each word.

\begin{table}[ht]
    \centering
    \scalebox{0.98}{
    \begin{tabular}{ccp{4.2cm}cccccc}
        \hline
        \bf  & \bf  & \bf  & \multicolumn{5}{c}{\bf Annotations} &   \\  
        \bf ID & \bf Corpus & \bf Context & \bf VE & \bf E & \bf N& \bf D & \bf VD & \bf S-W  \\\hline
        1 & Bible & You will have treasure in \textbf{heaven}. & 24 & 1 & 3 & & & 0.423 \\\hline
        2 & Europarl & \textbf{Election} of Vice-Presidents (first, second and third ballots) & 19 & 1 & 5 & & & 0.544 \\\hline
        3 & Biomed & Annexin A7 was isolated as the agent that mediated aggregation of chromaffin \textbf{granules} and fusion of \ldots & & 14 & 2 & 4 & & 0.612 \\\hline
        4 & Bible & Ehud made him a sword which had two edges, a \textbf{cubit} in length; and he wore it under his clothing on his right thigh. & 2 & 3 & 4 & 12 & 8 & 0.848 \\\hline
        5 & Europarl & I therefore wanted to tell you that I inadvertently voted "yes' in the vote on the Cornelissen report on the first part of \textbf{recital} 0, when I intended to vote "no' . & 1 & 11 & 3 & 9 & 2 & 0.848 \\\hline
        6 & Europarl & The Rospuda valley is the last \textbf{peat} bog system of its kind in Europe. & 2 & 11 & 3 & 4 & 4 & 0.848 \\ \hline
        7 & Biomed & Amyloid burden worsens significantly with age, and by 9 mo, the \textbf{hippocampus} and cortex of untreated mice are largely filled with aggregated peptide. & 1 & 8 & 7 & 9 & 2 & 0.901 \\\hline
    \end{tabular}
    }
    \caption{Examples of annotations with interesting distributions indicating disagreement among annotators respectively. The target word in the context is highlighted in bold. S-W stands for Shapiro-Wilk. Levels of Annotations are Very Easy (VE), Easy (E), Neutral (N), Difficult (D), and Very Difficult (D).}
    \label{tab:Shapiro_examples}
\end{table}

\subsection{CompLex 2.0 Features}  \label{sec:corpus_ftrs}

We have presented a corpus that was developed according to the recommendations that we have set out earlier in this work (see Section \ref{sec:building_on}). Whilst we have made every effort to follow these, practical concerns have led us to pragmatic design decisions that made the development of our corpus feasible. In the following list, we itemise the design decisions that were made during the construction of our corpus and show how these link to the recommendations from Table \ref{tab:cwi_spec}.

\begin{enumerate}
    \item \textbf{Continuous Annotations:} We have implemented this using a Likert Scale as described above. Unlike Maddela-2018 who used a 4-point Likert scale, we chose a 5-point Likert scale to allow annotators to give a neutral judgment. To give final complexity values we took the mean average of these annotations, transforming the complexity labels in the range 0--1.
    
    \item \textbf{Context:} We presented annotations in context to the annotators and explicitly asked annotators to judge a word based on its contextual usage (but not on the context itself). Following \cite{peirce-1906}, we distinguish between word types (the distinct words used in a text, which comprise its vocabulary) and word tokens (the different occurrences or instances of those words throughout the text). There are clear variations in the complexity of different tokens sharing the same word type. For example, the word `table' receives  a higher complexity rating in the less common sense of `table a motion' than in the more  frequent sense of something being `on the table'.
    
    \item \textbf{Multiple Tokens:} We presented a maximum of 5 tokens per word type, per genre. This led to 5,617 word types across 10,800 tokens and contexts giving an average density of 1.92 contexts per word type. Although some word types do appear in multiple contexts 3,423 words appear with only a single context. 671 word types feature 5 or more tokens (and contexts). This is a compromise between our desire to include a wide variety of word types in the dataset and to include multiple tokens of each type. A dataset featuring a more rigorous treatment of contexts may reveal the role of context in complexity estimation in a way that our data is not able to. 
    
    \item \textbf{Multiple Token Annotations:} We have described our process of gathering an average of 25.75 annotations per token. We could have  chosen to  do fewer annotations in favour of annotating more tokens, however we prioritised having a large number of judgments per token to give a more consistent and representative averaged annotation.
    
    \item \textbf{Diverse Annotators:} We did not place many restrictions, or record demographic information regarding our annotators. Doing so may have helped to better understand the makeup of our annotations and identify potential biases. We did not record this information due to the crowd-sourcing setting that we used. This is something for future LCP annotation efforts to consider.
    
    \item \textbf{Multiple Genres:} We have selected three diverse genres with a potential for complex language. We deliberately avoided the use of Wiki text as this has been studied widely already in CWI. 
    
    \item \textbf{Multi-word Expressions:} We have included these in a limited form as part of our corpus. The MWEs make up 16.66\% of our corpus. We have included these as an interesting area to study and we hope that their inclusion will shed light on the complexity of MWEs. Further studies could focus on specific types of MWE, extending our research.
    
\end{enumerate}

\noindent
The CompLex 2.0 corpus is designed according to the recommendations we have set out. In particular, we do not record demographic information on our participants and as such cannot make reasonable claims as to the diversity of our annotators.  Our corpus is intended as a starting point for future LCP researchers to build on. Using the methodology described in this section, further datasets encoding information about complex words can be annotated, focusing on the remaining open research questions in lexical complexity prediction.

\section{Predicting Categorical Complexity} \label{sec:analysis_of_existing}

%%%%%%%%%%%%%%%%%%
 % \begin{itemize}
 %    \item apply features from section before to the three datasets in question and present stats on these
  %   \item we can then use these findings to motivate discussions on positive and negative points of each dataset (see section below!)
 % \end{itemize}
%%%%%%%%%%%%%%%%%%%%%%%%%%%%%%%%%%%%%%%%

We represented words and multiword units in the CWI--2016 \cite{paetzold-specia:2016:SemEval1}, CWI--2018 \cite{yimam2018report}, and the new CompLex 2.0 single word and multiword datasets using features which, on the basis of previous work in lexical simplification \cite{paetzold-2016}, text readability \cite{yaneva-2017,deutsch-2020}, psycholinguistics/neuroscience \cite{yonelinas-2005}, and our inspection of the annotated data, we consider likely to be predictive of their complexity (Section \ref{sec:analysis_of_features}).

%\todo[]{Was I working with CompLex 1.0 or CompLex 2.0?}

We used the \texttt{trees.RandomForest} method distributed with Weka \cite{Hall2009} to build baseline lexical complexity prediction models exploiting the features presented in Section \ref{sec:analysis_of_features}. In the experiments described in the current Section, we framed the prediction as a classification task with continuous complexity scores mapped to a 5-point scale. The points on these scales denote the proportions of annotators who consider the word complex (c): few ($0 \leq c < 0.2$), some ($0.2 \leq c < 0.4$), half ($0.4 \leq c < 0.6$), most ($0.6 \leq c < 0.8$), and all ($0.8 \leq c \leq 1$). 

Table \ref{table:evaluationBaselineRandomForests} displays weighted average F-scores and mean absolute error (MAE) scores obtained by the baseline models in the ten-fold cross validation setting. This table includes statistics on the number of instances to be classified in each dataset.

%\todo[size=\small]{Commas as digit group separators. Some countries use full stops instead. I got into the habit of using half-spaces ($\backslash$,), which is in line with the SI/ISO 31-0 standard. For consistency, I have gone back to full commas in this paper.}
%%%%%%%%%%%%%%%%%%%%%%%%%%%%%%%%%%%%%%%%
\begin{table}[ht]
% \begin{footnotesize}
%\begin{small}
\begin{centering}
% \resizebox{\textwidth}{!}{
\begin{tabular}{cccc}
\hline
       & \bf F-score  & & \\
\bf  Dataset & \bf (weighted average) & \bf  MAE & \bf  Instances \\ \hline
CWI 2016 & 0.915 & 0.04 & $2,237$ \\
CWI 2018 & 0.843 & 0.0681 & $11,949$ \\
CompLex 2.0 (single) & 0.607 & 0.1782 & $7,233$ \\
CompLex 2.0 (MWE)  & 0.568 & 0.2137 & $1,465$ \\
\hline
\end{tabular} %}
\caption{Evaluation results of the baseline \texttt{trees.RandomForest} classifier. \label{table:evaluationBaselineRandomForests}}
\end{centering}
%\end{small}
\end{table}
%%%%%%%%%%%%%%%%%%%%%%%%%%%%%%%%%%%%%%%%

%\todo[]{Should we be consistent w.r.t ``multi words'' vs. ``multiple words''?}
Table \ref{table:featureAblation} displays the results of an ablation study performed in order to assess the contribution of various groups of features to the word complexity prediction task applied in the four datasets: CWI--2016, CWI--2018, CompLex (single words), and CompLex (multi-words). The feature sets refer to those studied previously in this work in Section \ref{sec:analysis_of_features}. In the table, negative values of  $\Delta$MAE indicate that the features are helpful, reducing the mean absolute error of the classifier. The reverse is true of positive values. 

%%%%%%%%%%%%%%%%%%%%%%%%%%%%%%%%%%%%%%%%
\begin{table}[ht]
% \begin{footnotesize}
\begin{small}
\begin{centering}
% \resizebox{\textwidth}{!}{
\begin{tabular}{cllll}
\hline
 & 	\multicolumn{4}{c}{\bf $\Delta$MAE} \\
\bf Ablated & 	 & 	 &  \bf	CompLex  & \bf	 	CompLex   \\
\bf	 feature group & \bf		CWI--2016 & \bf		CWI--2018 & \bf		 (single) & 	\bf	 (multi)  \\ \hline
% None & 	0 & 	0 & 	0 & 	0 \\
A       & 	+1E-04  & 	0       & 	+0.0002 & 	-0.0002 \\
B       & 	+0.0002 & 	+0.0002 & 	-1E-04  & 	-0.0006 \\
C       & 	-0.0001 & 	+0.0004 & 	+1E-04  & 	-0.0004 \\
D       & 	-0.0001 & 	+0.0001 & 	-1E-04  & 	-0.0002 \\
E, F, G & 	0       & 	+0.0001 & 	+0.0003 & 	+0.0006 \\
H, I, J & 	+0.0002 & 	+0.0004 & 	+0.0007 & 	+0.0012 \\
M       & 	-0.0001 & 	+0.0001 & 	0       & 	-0.001 \\
N       & 	0       & 	0       & 	+1E-04  & 	+0.0005 \\
P       & 	-0.0002 & 	+0.0001 & 	+0.0002 & 	-0.0007 \\
Q       & 	0       & 	+0.0001 & 	+1E-04  & 	-0.0002 \\
R       & 	+1E-04  & 	0       & 	-1E-04  & 	-0.0009 \\
S       & 	+1E-04  & 	+0.0001 & 	0       & 	-1E-04 \\
Linguistic features (A-S)  & 	0 & 	+0.0009 & 	\textbf{+0.0018} & 	\textbf{+0.0027} \\
%(A-S) & 	 &  & 	 & 	 \\
T & 	\textbf{-0.0029} & 	\textbf{+0.002} & 	\textbf{+0.001} & 	\textbf{-0.0065} \\
All but C  & 	\textbf{+0.0093} & 	\textbf{-0.0681} & 	\textbf{+0.0469} & 	\textbf{+0.0396} \\
\hline
% All but C (Naïve Bayes) & 	  & 	0.0323 & 	0.0422 & 	0.0323
\end{tabular} %}
\end{centering}
\caption{Results of feature ablation. Positive numbers represent a higher MAE after the named feature group was removed (hence the feature was helpful), whereas negative numbers represent the opposite. Most deltas are small, indicating minimal effect from many features. Values above 0.001 or below -0.001 are highlighted in bold.} \label{table:featureAblation}
\end{small}
\end{table}
%%%%%%%%%%%%%%%%%%%%%%%%%%%%%%%%%%%%%%%%

Our results indicate that for prediction of lexical complexity in the CWI--2016 dataset, five of the ablated feature groups are useful. Features encoding information about word length and the regularity of the singular/plural forms of nouns, the typical age of acquisition of the words, and the broad syntactic categories of the words improve the accuracy of the classifier, as do word embeddings.

For words in the CWI--2018 dataset, no feature group was found to be particularly useful for prediction of lexical complexity, though a simple model based only on word length information outperformed the default baseline exploiting all features. Again, this may be due to the typically longer MWEs present in the CWI--2018 dataset, which are exclusively labelled as complex.

When predicting the lexical complexity of individual words in the Comp\-Lex 2.0 data, features encoding information about whether or not the word was archaic, about the regularity of the singular/plural forms of nouns, and about the stress patterns of the words were all found to be useful. When considering multiword units (bigrams), a far larger proportion of the feature groups was observed to be useful for lexical complexity prediction. In our ablation study of bigrams, we assigned the bigram the average value of each feature (all of the features were represented numerically, including binary and one hot representations, and none of the features were symbolic). We found that features encoding information about word frequency, whether or not the words were archaic, word length, regularity of singular/plural forms, standard age of acquisition, broad syntactic category, the word's status as either archaic, alien, obsolete, colloquial, rare, or standard, the stress pattern of the word, and the occurrence of an INFOBOX element in the Wikipedia entry for the word were all useful predictors of lexical complexity. Averaged word embeddings also improved the accuracy of predictions made by the \texttt{trees.RandomForest} classifier in the CompLex (multi) dataset. 

%\todo[]{Forcing hyphenation here}
In the CWI--2016 and CWI--2018 datasets, we applied Weka's attribute (feature) ranking method with the unsupervised \texttt{Principal} \texttt{Components} Attribute Transformer evaluator to the 378 numerical features described previously in Tables \ref{table:wordFeaturesAJ} and \ref{table:wordFeaturesKT} (Section \ref{sec:analysis_of_features}). Table \ref{table:featureSelection} displays the ten top-ranked groups of features for the four datasets. The main observations to be drawn from the feature selection study is the usefulness of information related to word familiarity, concreteness, and imageability in all datasets and information from the vector representations of words derived using GloVe \cite{pennington2014glove}. These features occur in the systems that participated in the CWI Shared Tasks as shown in Tables \ref{tab:approaches} and \ref{tab:approaches2018}. This corroborates our findings in line with previous work.

\begin{table}[ht]
% \begin{footnotesize}
\begin{small}
\centering
% \resizebox{\textwidth}{!}{
\begin{tabular}{ccccc}
\hline
 &  & & \bf CompLex & \bf CompLex \\
\bf Rank & \bf CWI--2016 & \bf CWI--2018 & \bf (single) & \bf (multi) \\ \hline
1    & E, F, G, K & E, F, G &  E, F, G &   T (subset) \\
2    & T (subset) & E, F, G     & T (subset) &   T (subset) \\
3    & H, I, J , T (subset) &     & T (subset) &   E, F, G \\
4    & T (subset) & T (subset) & T (subset) &   T (subset) \\
5    & T (subset) & T (subset) & D, N, A &   T (subset) \\
6    & T (subset) & D, T (subset) & T (subset) &   T (subset) \\
7    & T (subset) & T (subset) & T (subset) &   T (subset) \\
8    & T (subset) & T (subset) & T (subset) &   T (subset) \\
9    & T (subset) & T (subset) & T (subset) &   T (subset) \\
10    & T (subset) & P, T (subset) & T (subset) &   T (subset) \\
\hline
\end{tabular} %}

\caption{Results of feature selection (\texttt{PrincipalComponents}).} \label{table:featureSelection}
\end{small}
\end{table}

Interestingly, whereas the results presented previously using a correlation analysis did not find psycholinguistic features (Groups E,F,G,K) to be correlative with complexity, the principal component analysis indicates that these features are in fact likely to be useful for prediction in these datasets.

These results demonstrate that by using our new data from CompLex 2.0, the features that we expect to correlate well with complexity judgments are more likely to be effective features for classification than when annotations are done in a binary setting  as in the CWI--2016 and CWI--2018 datasets.

\section{Predicting Continuous Complexity}
\label{sec:complex_experiments}

%\begin{itemize}
%    \item There's loads we can add in here, e.g., train on one genre predict on another, look at user-specific annotations, provide some baselines on the corpus (e.g., using the features from above, etc.)
%    \item it would also be great to analyse the features above and compare them to those on the other CWI datasets we're looking at.
%\end{itemize}

In our final section, we use the data we have collected to discuss the nature of complex words from a different perspective than in Section \ref{sec:analysis_of_existing}. Whereas in the previous Section we converted all labels into a categorical format to allow comparison, in this Section we use the labels assigned to CompLex 2.0 to discuss factors affecting the nature of lexical complexity, and its prediction. We first look at the effects of genre on CWI. We then continue in our exploration to study the distribution of annotations, to determine how and when annotators agree on the complexity of a word.

\subsection{Prediction of Complexity Across Genres}

%One feature of our corpus is that it includes texts of several genres. We have included diverse genres as we wish to ensure that models derived from our data are transferable to domains outside of those in our corpus. This is important for CWI as previous corpora have focused on a restricted range of genres (encyclopaedic text from Wikipedia and news text) and domains, which may limit their applicability beyond that restricted set.

%\todo[]{Can we be more explicit about our goal here? Why are we doing this?} 
To better understand the effect of text genre on the LCP task we designed the experiments described in this Section. For these, we employed a simple linear regression with the features described previously in Section~\ref{sec:analysis_of_features}. 
We use the single words in the corpus and split the data into training and test portions, with 90\% of the data in the training portion and 10\% of the data in the test portion. We first created our linear regression using all the available training data and evaluated this using Pearson's Correlation. We used the labels given to the data during the annotation round we undertook to create CompLex 2.0. The prediction model based on linear regression achieved a score of 0.771, indicating a reasonably high level of correlation between its predictions and the labels of the test set.   

This result is recorded in Table~\ref{tab:predicting}, where we also show the results for each genre. In each case, we have selected only data from a given genre and followed the same procedure as above, splitting into train and test and evaluating using Pearson's correlation. The linear regression model is less closely correlated when making lexical complexity predictions in the  Europarl (0.724) and in the Bible data (0.735). This is expected, given the reduction in size of the training data.  It is surprising to see that the linear regression model worked better for the Biomedical data than for any other subset  (0.784). This may indicate that simple and complex words are more distinct in this corpus and that this distinction can be learnt from a more focused training set.

\begin{table}[ht]
    \centering
    \begin{tabular}{cc}
    \hline
       \bf  Subset   & \bf Correlation  \\\hline
         All      & 0.771 \\
         Europarl & 0.724 \\
         Biomed   & 0.784\\
         Bible    & 0.735 \\
         \hline
    \end{tabular}
    \caption{Results of training a linear regression on all the data, and on each genre.}
    \label{tab:predicting}
\end{table}

%\subsection{Genre Experiments}
%Train on 1 genre - test on other 2.
%Train on 2 genres - test on other 1.

To  further determine the effects of genre on lexical complexity prediction, we constructed a new linear regression model that was trained and tested using specific genres selected from our corpus. We trained on single genres and tested on each of the other 2 genres, as well as training on a combined subset of 2 genres and testing on the remaining genre. The results for this experiment are shown in Table \ref{tab:genre}. We were able to build a reliable predictive model for cross-genre complexity prediction in each case.

Our results show that there is a drop in performance when training on out-of-domain data, compared to training on in-domain data. This is true across all genres, where a reduction of between 0.119 and 0.297 can be observed in Pearson's correlation. In each genre, the scores improve when training on the other two genres, rather than just on one.  This may  be  due to the effect of multiple genres helping the linear regression to generalise to global  complexity effects, rather  than overfitting to  specific complexity features in one genre. If we were to  test our results on an additional genre/domain, we  may hope to see that training on three genres  (as  are present in our  corpus) would yield even more generalised results.

\begin{table}[ht]
    \centering
    \begin{tabular}{ccc}
        \hline
          \bf Train & \bf Test & \bf Correlation  \\\hline
         Biomed   & Europarl & 0.542 \\
         Bible    & Europarl & 0.484 \\
         Biomed + Bible & Europarl & 0.651 \\\hline
         Bible    & Biomed   & 0.487 \\
         Europarl & Biomed   & 0.630 \\
         Bible + Europarl & Biomed & 0.723 \\\hline
         Biomed   & Bible    & 0.605 \\
         Europarl & Bible    & 0.616 \\
         Biomed + Europarl & Bible & 0.692 \\
         \hline
    \end{tabular}
    \caption{Results of training a linear regression on one genre, or pair of genres and testing on a different genre.}
    \label{tab:genre}
\end{table}

\subsection{Subjectivity}

%Words which annotators generally agreed upon
%Words which annotators generally disagreed upon
%Can we predict how subjective a word is?

We previously used a Shapiro-Wilk test to demonstrate that our annotations are generally normally distributed. We obtained the mean of each annotation distribution to give a complexity score for each instance in our dataset. An interesting question to ask is how representative these means are of the true complexity of a word. One word may be considered easy by one annotator, yet difficult by another. Factors such as age, education and background may well affect which words a reader is familiar with. We can use the normally distributed annotations to understand this phenomenon by investigating the standard deviations of the annotations for each instance.

We have provided examples from our corpus in Table \ref{tab:stdev_examples} with both the mean complexity and the standard deviation ($\sigma$) of the annotations. The top three rows show examples of high standard deviation, whereas the bottom three rows show examples of low standard deviations.  It is clear from the table that annotators generally agree more about words which are less complex, with disagreements tending to happen around the more difficult words. An analysis of the mean complexity and standard deviation of the complexity yields a Pearson's correlation of 0.621, indicating that these are moderately correlated (disagreement is linearly related to complexity).

\begin{table}[ht]
    \centering
    \begin{tabular}{cp{5cm}cc}
        \hline
        \bf Corpus & \bf Context & \bf Complexity & \bf $\sigma$  \\\hline
        Biomed &  The first step requires generating a floxed allele in ES cells that will serve as the \textbf{substrate} for subsequent exchanges (RMCE-ready ES cell, Figure 1). & 0.556 & 0.433 \\
        Bible & The second came, saying, 'Your mina, Lord, has made five \textbf{minas}. & 0.433 & 0.423  \\
        Europarl & 'Budget support' refers to the transfer of financial resources from a funding agency outside the partner country's treasury, under the \textbf{proviso} that the country abide by the agreed conditions governing payments. & 0.567 & 0.382  \\
        Biomed & Similarly, changes in \textbf{synaptic plasticity} due to Ca2+-permeable AMPARs [51,52,60], e.g., in piriform cortex, might alter odor memorization processes. & 0.975 & 0.077  \\
        Bible & Or were you baptized into the \textbf{name} of Paul? & 0.000 & 0.000  \\
        Europarl & Therefore, I would like to ask, in accordance with the Rules of Procedure, for the \textbf{matter} to be referred to the competent body. & 0.175 & 0.118   \\
        \hline
    \end{tabular}
    \caption{Examples of instances with subjective (wide standard  deviation) and certain (narrow standard deviation) annotations.}
    \label{tab:stdev_examples}
\end{table}

\section{Discussion}\label{sec:discussion}

% introduction of new definition of LC

Our work has sought to introduce a new definition of lexical complexity to the research community. Whereas previous treatments of lexical  complexity have considered it a binary phenomenon in the Complex Word Identification (CWI) setting, we have extended this definition to lexical complexity prediction (LCP), considering complexity as a continuous value associated with a word.  This new definition asks the question of `how complex is a word' rather than `is this word complex or not?'. This question allows us to give each token a complexity rating on a continuous scale, rather than a binary judgment. If binary judgments were required, it would be easy to create them using our dataset by imposing a threshold at some point in the data. By imposing thresholds at different points,  binary labels can be obtained to suit different subjective definitions of complexity.  Further, by implementing multiple thresholds, multiple categorical labels can be recovered from the data.

% analysis of ftrs of complex words

In  Section \ref{sec:analysis_of_features} we showed that the types of features we would typically expect to correlate with word complexity did not show any correlation with the CWI--2016 and CWI--2018 datasets. This motivated our analysis of the protocol underlying the annotation of these datasets and our development of a new protocol for CWI annotation. In Section \ref{sec:analysis_of_existing}, we were able to show through the use of feature ablation experiments that more of the feature sets that we used were relevant to the classification of CompLex 2.0, than were relevant to the annotation of CWI--2016 or CWI--2018. This implies that the annotations in our new dataset are more reflective of traditional measures of complexity.

% specification for LC Datasets
%\todo[]{I'm not sure which section numbers you want to reference here.}
We discussed the existing CWI datasets at length (Section~\ref{sec:analysis_of_features}), culminating in our new specification for LCP datasets in Section~\ref{sec:spec_cwi}. Whilst we have gone on to develop our own dataset (CompLex 2.0), we also hope to see future work developing new CWI datasets following the principles that we have laid out. 
Future datasets could focus on multilinguality, multi-word expressions, further genres, or simply extending our analysis to further tokens and contexts.  
Certainly, we do not see the production of CompLex 2.0 as an end point in LCP research, but rather a starting point for other researchers to build from. 
This is why we have included our protocol in detail --- in  order to ensure the replicability of our work in future research. 

In moving from binary annotations to Likert-scale annotations, we have provided a new dataset, which gives continuous annotations based on a more objective measure of complexity. The binary setting could also be improved if more objective guidelines were provided to the annotators (e.g., instructions such as ``identify words that are appropriate for an adult", or ``identify words that are specific to a domain", as opposed to ``identify words that \textbf{you} find difficult). In our comparison, we are comparing a subjective binary dataset to a (more) objective continuous dataset (of course, our dataset still relies on some degree of annotator interpretation of the Likert scale labels). We do not have the ability to compare an objective binary dataset to our data, as it does not exist to the best of the author's knowledge, however doing so would likely yield further interesting insights into the differences between continuous and binary lexical complexity.

% CompLex 2.0 development

We implemented our specification for a new LCP dataset, following the recommendations established in Section \ref{sec:analysis_of_features}. This led to the creation of CompLex 2.0. In Section \ref{sec:corpus_ftrs} we have explicitly compared our dataset to the recommendations  we made in Table~\ref{tab:cwi_spec}, and we would encourage the creators of future LCP datasets to do the same. This will ensure that datasets can be  easily evaluated  and compared at a feature level. The CompLex 2.0 dataset is available via GitHub\footnote{\url{https://github.com/MMU-TDMLab/CompLex}}. We have made this data available under a CC-BY licence, facilitating its reuse and reproducibility outside of our work. 

Our new LCP dataset is the first to provide continuous complexity annotations for words in context. The role of context in lexical complexity has not been widely studied and we  hope that  this dataset will go some way towards allowing researchers to work on this topic. Indeed, the evidence from our annotations shows that for a single token in multiple contexts, the complexity annotation of that token does vary. Further work is needed to prove that the variation is an effect of the contextual occurrence, or difference in sense and not due to the stochastic nature of annotations resulting from crowdsourcing.

% Soundness of moving from categorical labels to continuous labels
Although we gave annotators in our task a 5-point scale ranging from Very Easy to Very Difficult, we chose to aggregate the annotations to give a mean-average for each instance. This makes a fundamental  assumption that the distance in  continuous complexity space between each point on the Likert scale is constant. Obviously, there is  no guarantee that such an assumption is true. The danger of this is that annotations may be falsely biased towards one end of the scale. For instance, if the distance between Very Easy and Easy is shorter than the distance between Easy and Neutral, then considering these as the same distance will falsely inflate complexity ratings. Another strategy could have been to take the median or mode of the complexity annotations to give a final value. The disadvantage of that approach would be that every instance would have an ordinal categorical label instead of a continuous label as we have advocated for.  This would be a different problem to the one we have explored, and is left to future research.

% Table 13

We used categorical complexity to provide a feature analysis of our dataset and the prior CWI datasets in Section \ref{sec:analysis_of_existing}. We observed that a number of features were identified as useful for the prediction task, indicating that complexity is a matter of many factors and no single factor can be used  to determine a word's complexity. Interestingly, in Table \ref{table:evaluationBaselineRandomForests}, we  showed that in the categorical setting the CWI-2018 and CWI-2016 datasets both outperformed the CompLex 2.0 dataset. We are not trying to use the dataset here to demonstrate some superior performance, but rather demonstrate a comparative analysis of features that are useful for complexity prediction. This may indicate that systems wishing to return a categorical label (as those used in Section \ref{sec:analysis_of_existing}), could use probabilistic or categorical data for training and get better results than when using our data. Our continuous labels allow us to  perform further interesting analyses into the nature of complex words as presented in Section \ref{sec:complex_experiments}.

% SemEval 2021 ST

%We used the data from CompLex 2.0 to run a shared task at SemEval in 2021 \cite{}. We split the data into trial, train, and test partitions, putting 10\% of the data into each of trial and test and the remaining 80\% into the training set. We stratified the data according to complexity, as well as ensuring that there were not tokens shared between datasets. We released the trial and training data with labels for participants to build systems with, before releasing the test data without labels and asking participants to submit their predictions.  We analysed the predictions and returned scores to the participants. The best scoring systems were able to attain in excess of 0.78 Pearson's correlation for the single word data and in excess of 0.86 Pearson's correlation for the combined single word and multi word data.

% experiments with CompLex 2.0
We were able to use our data to show that complexity can be predicted  across genres. This is encouraging as our dataset contains three diverse genres, and we can expect that the complexity annotations we have identified will generalise well to other genres. A model trained on all three genres will learn features of complexity that are common to all genres, rather than to any one specific genre. We also demonstrated that our instances vary in subjectivity of complexity, with those rated as more complex typically  being more subjective. Identifying the factors that make a  word subjectively complex would be an interesting line of study, but is left for future research.

% complexity and ambiguity
The ambiguity of a word is likely to play a role in its complexity. Words which are often mistaken for others are more likely to be confused and hence are likely to be rated as more difficult to understand by a reader. Conversely however, there is a well documented direct correlation between polysemy and frequency (i.e., infrequent words are typically monosemes, whereas frequent words have many senses. See the WordNet entries for `run', `bat', `cat', etc.). It may be hypothesised that ambiguity and frequency need to be jointly taken into account when investigating lexical complexity, with a likely ordering from least to most complex being: (high-frequency, monosemous), (high-frequency, polysemous), (low-frequency, monosemous), (low-frequency, polysemous). Prior efforts have been  undertaken to create sense annotated complexity datasets \cite{strohmaier-etal-2020-secoda}, and building upon our research with sense annotations, using the specification given in Section \ref{sec:specification} will lead to fruitful research outcomes.

\section{Conclusion}\label{sec:conclusion}

We have demonstrated that previous datasets are insufficient for the task of Complex Word Identification. In fact, the very definition of the task --- identifying complex words in a subjective binary setting rather than on an objective continuous scale is at fault. We have advocated for a generalisation of this task to Lexical Complexity Prediction and we have provided recommendations for datasets approaching this task. Further to this we have provided a new dataset, CompLex 2.0, which is the first publicly available dataset to provide continuous complexity annotations for words in context. We release the data in full to allow future researchers to join us in this exciting task of Lexical Complexity Prediction.
%\todo[]{Can this be worded more formally?}

%\begin{acknowledgements}
%\end{acknowledgements}

\bibliographystyle{spmpsci}      % mathematics and physical sciences
\bibliography{Thesis,bibliography_Matt,bibliographymarcos,bibliographyEvans}   % name your BibTeX data base

\end{document}